\documentclass[lettersize,journal]{IEEEtran}
\usepackage{amsmath,amsfonts}
\usepackage{algorithmic}
\usepackage{algorithm}
\usepackage{array}
\usepackage[caption=false,font=normalsize,labelfont=sf,textfont=sf]{subfig}
\usepackage{textcomp}
\usepackage{stfloats}
\usepackage{url}
\usepackage{verbatim}
\usepackage{graphicx}
\usepackage{cite}
\usepackage{makecell}   
\usepackage{dsfont}     
\usepackage{enumitem}
\usepackage{booktabs} 
\usepackage{multirow} 
\usepackage{nicematrix} 
\usepackage{amssymb} 
\usepackage[table,xcdraw]{xcolor}
\usepackage[table]{xcolor}
\usepackage[colorinlistoftodos]{todonotes}
\usepackage{soul}  

\usepackage[colorlinks,
            linkcolor=blue,
            anchorcolor=blue,
            citecolor=blue]{hyperref}

\begin{document}

\title{SMART: Semantic Matching Contrastive Learning for Partially View-Aligned Clustering}


\author{
Liang Peng$^\dagger$,
Yixuan Ye$^\dagger$,
Cheng Liu$^{*}$, \textit{Senior Member, IEEE},
Hangjun Che, \textit{Senior Member, IEEE},
Fei Wang,
Zhiwen Yu, \textit{Senior Member, IEEE},
Si Wu,
and~Hau-San Wong
\thanks{Cheng Liu is with the College of Computer Science and Technology, Huaqiao University and the Department of Computer Science, Shantou University (email: chengliu10@gmail.com)}
\thanks{Liang Peng, Yixuan Ye, Fei Wang are with the Department of Computer Science, Shantou University (email: 23lpeng@stu.edu.cn; 22yxye2@stu.edu.cn;  wangfei@stu.edu.cn)}
\thanks{Hangjun Che is with the Chongqing Key Laboratory of Nonlinear Circuits and Intelligent Information Processing, College of Electronic and Information Engineering, Southwest University, Chongqing, China (email: hjche123@swu.edu.cn)}
\thanks{Zhiwen Yu and Si Wu are with the School of Computer Science and Engineering, South China University of Technology. (email: zhwyu@scut.edu.cn, cswusi@scut.edu.cn)}
\thanks{Hau-San Wong is with the Department of Computer Science, City University of Hong Kong. (email: cshswong@cityu.edu.hk)}
\thanks{$\dagger$ These authors contributed equally and are co-first authors.}
\thanks{$^{*}$ Corresponding author: Cheng Liu}
}



\maketitle

\begin{abstract}
Multi-view clustering has been empirically shown to improve learning performance by leveraging the inherent complementary information across multiple views of data. However, in real-world scenarios, collecting strictly aligned views is challenging, and learning from both aligned and unaligned data becomes a more practical solution. Partially View-aligned Clustering (PVC) aims to learn correspondences between misaligned view samples to better exploit the potential consistency and complementarity across views, including both aligned and unaligned data.
However, most existing PVC methods fail to leverage unaligned data to capture the shared semantics among samples from the same cluster. Moreover, the inherent heterogeneity of multi-view data induces distributional shifts in representations, leading to inaccuracies in establishing meaningful correspondences between cross-view latent features and, consequently, impairing learning effectiveness.
To address these challenges, we propose a \textbf{S}emantic \textbf{MA}tching cont\textbf{R}as\textbf{T}ive learning model (SMART) for PVC. The main idea of our approach is to alleviate the influence of cross-view distributional shifts, thereby facilitating semantic matching contrastive learning to fully exploit semantic relationships in both aligned and unaligned data. Specifically, we mitigate view distribution shifts by aligning cross-view covariance matrices, which enables the inference of a semantic graph for all data. Guided by the learned semantic graph, we further exploit semantic consistency across views through semantic matching contrastive learning. 
After the optimization of the above mechanisms, our model smoothly performs semantic matching for different view embeddings instead of the cumbersome view realignment, which enables the learned representations to enjoy richer category-level semantics and stronger robustness.
Extensive experiments on eight benchmark datasets demonstrate that our method consistently outperforms existing approaches on the PVC problem.
The code is available at \url{https://github.com/THPengL/SMART}
\end{abstract}


\section{Introduction}
\begin{figure}[t]
    \centering
    \includegraphics[width=1\linewidth]{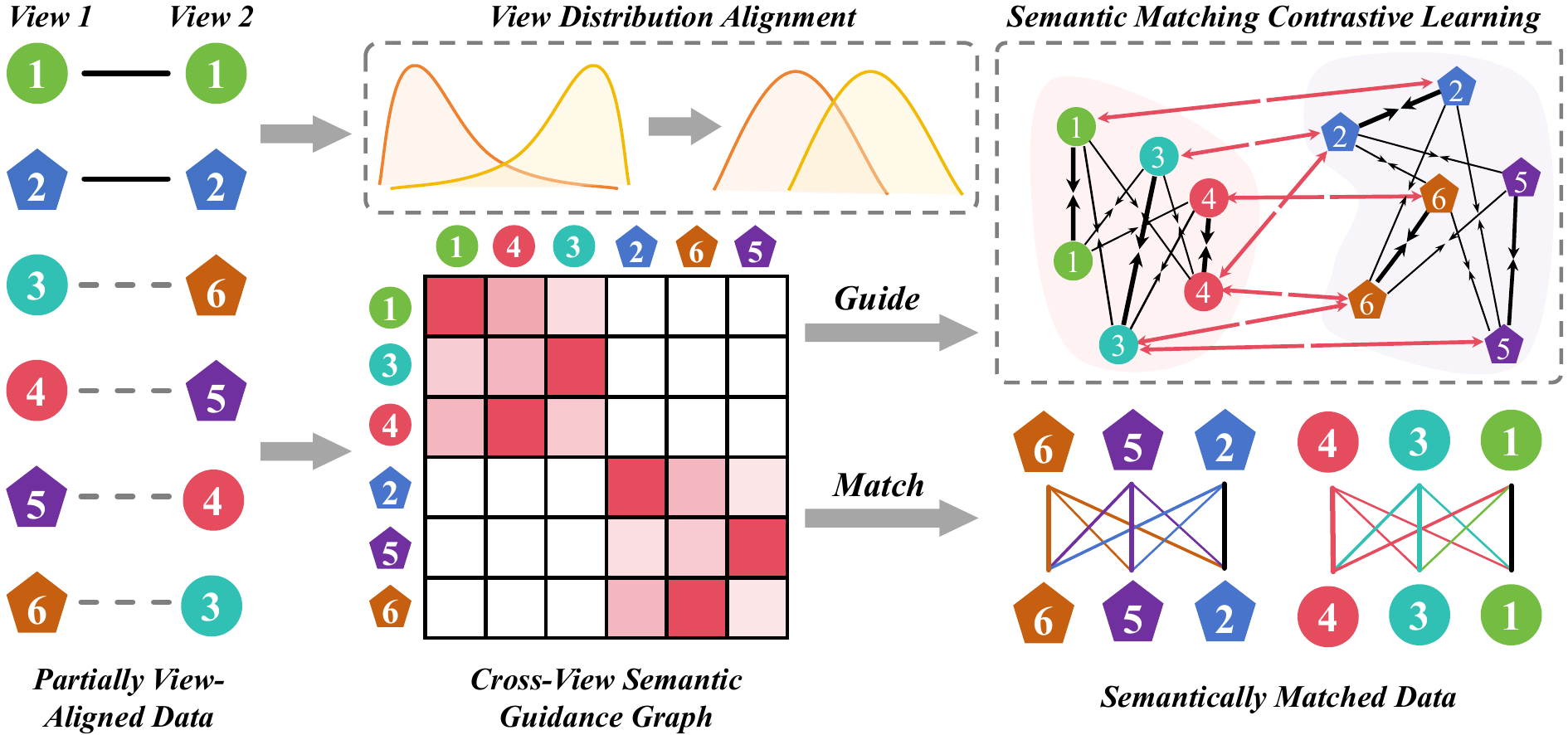}
    \caption{An example illustrates PVC setting and our model: different categories are represented by distinct shapes, and individual instances are characterized by different colors. Solid lines indicate aligned instances, while dashed lines represent unaligned ones. Black arrows indicate attraction between semantic pairs (weighted positive pairs), while red arrows indicate repulsion between negative pairs. Our method first performs view distribution alignment through aligned samples, then constructs a reliable semantic graph with all instances to guide the semantic matching contrastive learning. This ensures the model to capture common and complementary semantics from both aligned and unaligned data, thereby generating semantically matched meaningful and robust representations.}
    \label{fig:Motivation}
    \vspace{-1.0em}
\end{figure}
\IEEEPARstart{D}{iverse} real-world data can be represented through multiple heterogeneous feature representations derived from distinct modalities, such as images, text, and videos \cite{2024learnable-zhao, 2024evaluate-wang, 2024learning-zhang, 2024Differentiable-yan, 2024effective-hu-r5, 2023comprehensive-fang}. These heterogeneous representations are collectively referred to as multi-view data \cite{2023cross-dong-r5, 2025CVNC-ye, 2025DeepSurvey-Ren, 2024high-hu-r5}. Multi-view clustering (MVC) has demonstrated empirical success in enhancing clustering performance by leveraging the inherent complementary characteristics and shared information across different views \cite{2024Projective-Deng, 2024CTGL-liu, 2024Reliable-Hu, 2022learning-r1, 2023scDFC-hu-r5, 2024scEGG-hu-r5, 2020surface-cui-r5, xian2025mdhgfn, 2023SMILE-zeng-r1}.
However, existing MVC methods often rely on the idealized assumption that all views are perfectly aligned \cite{2024Dual-Optimized-Wen, 2024Automatic-Ma, 2024Dual-Cui, 2024Dynamic-Ren, 2023information-liu}. In reality, this assumption rarely holds, as real-world multi-view data frequently suffers from misalignment \cite{2020PVC-huang, 2021MvCLN-yang} or incomplete correspondences between views \cite{2020adaptive-wen, 2024Liu-sample, 2024adaptive-pu, 2024LaSA-liu}. This challenge gives rise to partially view-aligned clustering (PVC), which aims to effectively utilize both aligned and unaligned data \cite{2024CGDA-wang, 2024VITAL-he, 2024CGCN-wang, 2024dynamic-zhao}.
A fundamental challenge in PVC is to effectively exploit the limited consistent and complementary information available in partially aligned multi-view data to accurately capture the semantic relationships between samples from different views. 

To tackle the challenges of PVC, a common strategy is to learn correspondences between misaligned sample views to better exploit the potential consistency and complementarity across views, including both aligned and unaligned data. For instance, PVC \cite{2020PVC-huang} establishes correspondences between misaligned view samples in a common space using the Hungarian \cite{1955hungarian-kuhn} algorithm, while CGDA \cite{2024CGDA-wang} learns such correspondences through graph distributions that capture semantically invariant instance relationships across views. The above-mentioned methods primarily rely on instance-level alignment, aiming to accurately estimate the true relationships between unaligned samples across views. However, this strict instance-level correspondence learning can lead to unsatisfactory performance when the sample alignments are inaccurate.
Category-level alignment offers a way to alleviate the need for strict correspondences \cite{2024VITAL-he}, but it often heavily depends on known correspondences and one-to-one view realignment. Nevertheless, such approaches still fail to fully exploit the inherent semantic relationships (i.e., samples belonging to the same cluster across views should exhibit strong consistency) present in both aligned and unaligned data. For example, SURE \cite{2022SURE-yang}, a category-level alignment method, constructs sample pairs for contrastive learning to discover category-level correspondences but still relies completely on existing correspondences (i.e. aligned samples) for category-level view realignment. As a result, the potential of unaligned data in semantic learning has not been explored. Moreover, the inherent heterogeneity of multi-view data induces distributional shifts in representations and hinders the effective use of semantic information in unaligned samples, resulting in inaccuracies when establishing meaningful cross-view features correspondences. 

To address these challenges, we propose a \textbf{S}emantic \textbf{MA}tching cont\textbf{R}as\textbf{T}ive learning model (SMART) for PVC. The main idea behind our approach is to alleviate the impact of distributional shifts caused by the inherent heterogeneity of multi-view data, while simultaneously facilitating semantic matching contrastive learning for both aligned and unaligned data. Specifically, SMART comprises two key modules: view distribution alignment and semantic matching contrastive learning. As shown in Fig. \ref{fig:Motivation}, the view distribution alignment module mitigates inter-view distribution shifts via employing cross-view covariance matching alignment. By this way, it reduces the distributional gap across views and facilitates the inference of a cross-view semantic graph for both aligned and unaligned data.
Guided by the learned semantic graph, the semantic matching contrastive learning module further enhances semantic consistency within latent clusters through adaptive graph contrastive learning, effectively uncovering reliable semantic relationships across views for both aligned and unaligned samples.
Ultimately, through reliable semantic matching, the proposed approach is free of correspondence learning and obtains more robust representation capability.

The main contributions of our proposed SMART can be summarized as follows: 
\begin{itemize}
\item We propose a novel framework that adopts a feature distributional alignment perspective to effectively mitigate inter-view discrepancies by matching inter-view covariance statistics.
\item We leverage both aligned and unaligned instances through semantic matching contrastive learning to capture reliable, complementary, and consistent semantics, fundamentally distinguishing our approach from traditional view correspondence learning methods.
\item Extensive experiments conducted on several benchmark datasets in partially and fully aligned scenarios demonstrate the effectiveness and superiority of our method.
\end{itemize}

\section{Related Works}
Multi-view clustering (MVC) has experimentally demonstrated remarkable performance by leveraging the consistency and complementary structures across multiple views \cite{2019GMC-wang, 2021reconsidering-trosten, 2023contrastive-r1, 2023multi-zhao, 2024survey-evaluation-zhou, 2024learn-liu, 2024Robust-liu, 2024learnable-zhao, 2024decoupled-r1,peng2025refinementcontrastivelearningcellgene}. However, this ideal assumption often fails to hold, leading to the challenges of Incomplete Multi-view Clustering (IMVC) \cite{2023Self-SGC-Self, 2023scalable-wen, 2023incomplete-r1, 2023graph-wen}, Noisy Correspondence (NC) \cite{2024robust-sun} and Partially View-aligned Clustering (PVC) \cite{2020PVC-huang, 2021MvCLN-yang}. 
To address the IMVC and NC issues, numerous methods have been proposed to overcome issues related to missing views \cite{2024Low-Rank-Cui, 2024spectral-chen, 2024DVSAI-Yu, 2024diffusion-wen, 2024PGP-Self, 2025deep-jin-r5, 2025imputation-dai-r5} and noisy inter-view correspondences \cite{2024robust-guo-r1}, respectively.
Unlike the above-mentioned methods, PVC aim to fully utilize aligned samples to infer cross-view correlations for unaligned instances, followed by realignment to facilitate inference across the entire dataset \cite{2024CGDA-wang, 2022SURE-yang, 2024CGCN-wang, 2024dynamic-zhao, 2024VITAL-he}.

The pioneer PVC aims to learn potential common representations and a differentiable bipartite graph matching algorithm for instance-level alignment to establish relationships between unaligned views \cite{2020PVC-huang}. Subsequently, CGDA advances this paradigm by aligning graph distributions that encapsulate view-invariant instance relationships, thereby establishing cross-view instance correspondences \cite{2024CGDA-wang}. 
These correspondence learning methods rely on instance-level alignment, seeking to accurately estimate the true relationships between unaligned samples across views.

Moreover, MvCLN moves towards category-level alignment by constructing positive pairs from aligned instances and employing random sampling to generate negative pairs. It further mitigates the issue of false-negative pairs through a noise-robust contrastive loss, enhancing the robustness of view correspondence learning \cite{2021MvCLN-yang}. Extending this framework, SURE proposes an integrated approach that simultaneously addresses both IMVC and PVC, thereby enhancing its applicability to partially aligned and incomplete multi-view datasets \cite{2022SURE-yang}. More recently, CGCN introduces cross-view graph contrastive learning module based on intra-cluster and
inter-view consistency to perform cluster-level alignment and learn latent representation simultaneously \cite{2024CGCN-wang}. VITAL designs a variational inference framework that models Gaussian distributions in the latent space to facilitate the reconstruction of category-level inter-view relationships \cite{2024VITAL-he}. Additionally, TCLPVC develops a triple-consistency-driven representation fusion method to mitigate the influences of noise and complementary information on view alignment and clustering \cite{2025TCLPVC-gao}.
\begin{figure*}[t]
    \centering
    \includegraphics[width=1\textwidth]{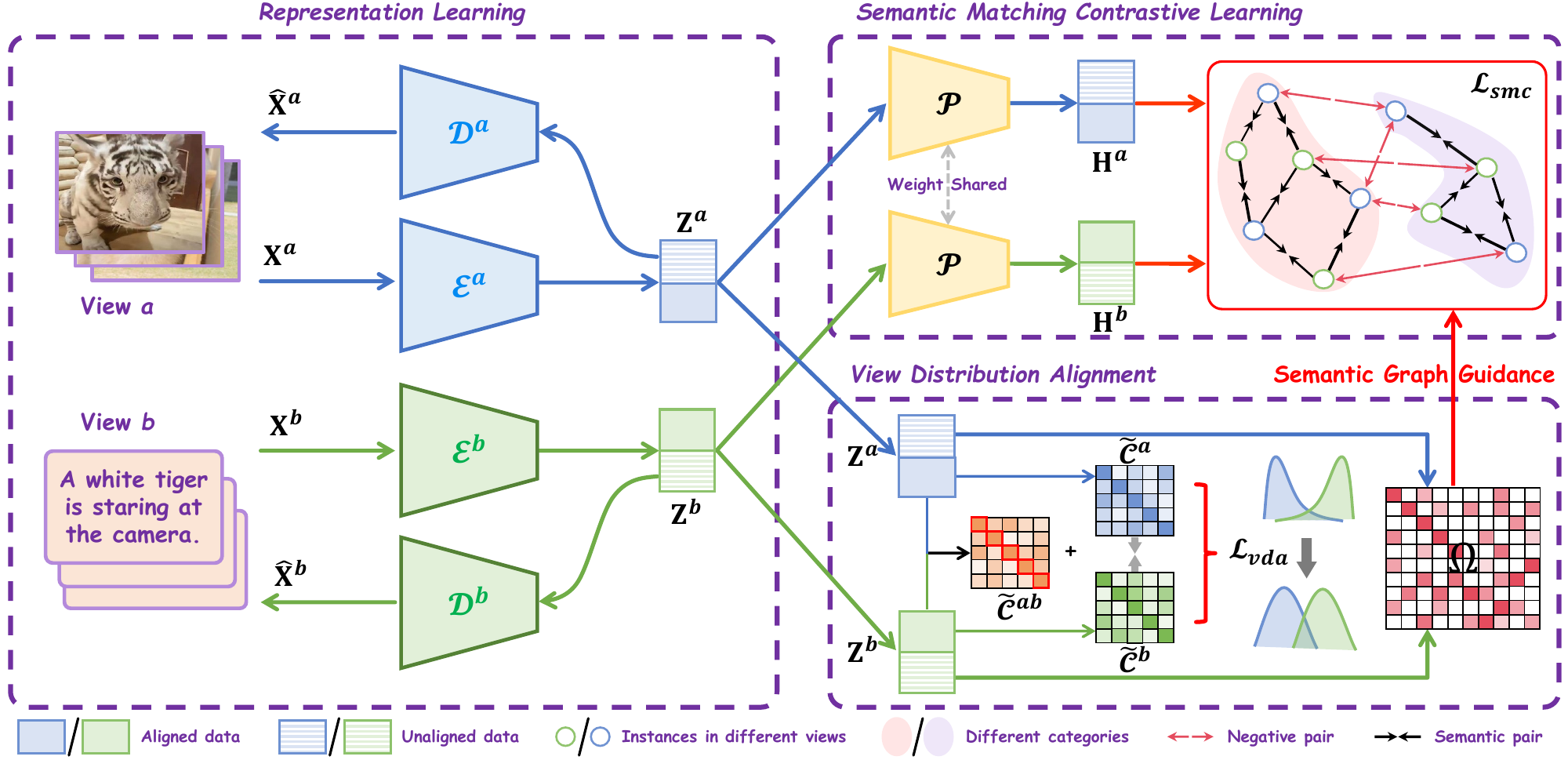}
    \caption{Overview of the proposed SMART method. Without loss of generality, we consider the two-view case. The view distribution alignment module jointly aligns cross-view covariance and minimizes discrepancies between view-specific covariance matrices, reducing inter-view distribution gaps. Meanwhile, a semantic guidance graph $\mathbf{\Omega}$ (as visualized in Fig. \ref{fig:adj}) is learned to capture complementary and consistent semantics from both aligned and unaligned data for semantic matching contrastive learning, thereby improving the representation learning capability. 
    }
    \label{fig:overview}
\end{figure*}
\begin{table}[!b]
\centering
\caption{Summary of notations in the main paper}
    \resizebox{0.5\textwidth}{!}{
    \renewcommand{\arraystretch}{1.0} 
    \begin{tabular}{l|l}
        \toprule[1.25pt]
        Notation & Description \\
        \midrule
        $V$ & Number of views. \\
        $N/N_a/N_u$ & Number of all/view-aligned/unaligned samples. \\
        $D_v/d$ & Dimension of the $v$-th view/latent embeddings. \\
        $\mathbf{X}^v \in \mathbb{R}^{N \times D_v}$ & Input features of the $v$-th view. \\
        $\mathbf{\widehat{X}}^v \in \mathbb{R}^{N \times D_v}$ & Reconstructed features of the $v$-th view. \\
        $\mathbf{Z}^v \in \mathbb{R}^{N \times d}$ & Latent embeddings of the $v$-th view.\\
        $\mathcal{E}^v/\mathcal{D}^v$ & Encoder/decoder of the $v$-th view. \\
        $\mathrm{cov}(\cdot)$ & Covariance operation. \\
        $\mu(\cdot)/\sigma(\cdot)$ & Mean value/standard deviation. \\
        $\mathrm{diag}(\cdot)$ & Row vector containing the diagonal elements. \\
        $\mathbf{\mathcal{C}}^{ab} \in \mathbb{R}^{N \times N}$ & Cross-view covariance matrix of all samples. \\
        $\mathbf{\mathcal{C}}^{v} \in \mathbb{R}^{N \times N}$ & Intra-view covariance matrix of all samples. \\
        $\widetilde{\mathbf{\mathcal{C}}}^{ab} \in \mathbb{R}^{N_a \times N_a}$ & Cross-view covariance matrix of aligned samples. \\
        $\widetilde{\mathbf{\mathcal{C}}}^{v} \in \mathbb{R}^{N_a \times N_a}$ & Intra-view covariance matrix of aligned samples. \\
        $\mathcal{P}$ & High-level projector. \\
        $\mathbf{H}^v \in \mathbb{R}^{N \times d}$ & High-level features of the $v$-th view.\\
        $\mathbf{\Omega}^{ab} \in \mathbb{R}^{N \times N}$ & Semantic guidance graph across views $a$ and $b$. \\
        $\mathcal{T}$ & The adaptive threshold to filter semantic pairs. \\
        $s(\cdot)$ & Cosine similarity. \\
        $\mathrm{norm}(\cdot)$ & $l_{2}$-normalization. \\
        $\mathrm{cat}(\cdot)$ & Concatenation operation. \\
        \bottomrule[1.25pt]
    \end{tabular}
    }
    \label{tab:notaions}
\end{table}
In contrast to existing PVC methods that focus on learning view correspondences for unaligned data, this work copes with the challenge from a contrastive learning perspective with reliable semantic matching. To be specific, we address a key limitation of existing PVC approaches, which often neglect the inherent heterogeneity of multi-view data, leading to distribution shifts and erroneous realignment that hinder the effective utilization of unaligned instances. The proposed SMART improves PVC tasks by integrating view distribution alignment and semantic matching contrastive learning, thereby reducing the distribution gap between heterogeneous views, avoiding cumbersome correspondence learning, and enhancing the capability to capture informative category-level semantics.

\section{Methodology}
\subsection{Preliminaries}
A partially view-aligned dataset can be formalized as $\{\textbf{X}^v\}_{v=1}^V$, where $V$ represents the total number of views. Each view consists of $N$ samples, denoted as $\textbf{X}^v = [\textbf{x}^v_1, \textbf{x}^v_2, \dots, \textbf{x}^v_N] \in \mathbb{R}^{N \times D_v}$, with $D_v$ indicating the feature dimensionality specific to the $v$-th view. The dataset is further divided into two subsets: aligned and unaligned data. Let $N_a$ and $N_u$ denote the numbers of aligned and unaligned samples, respectively, $N = N_a + N_u$.
Without loss of generality and following existing PVC methods \cite{2020PVC-huang, 2021MvCLN-yang, 2022SURE-yang, 2024CGDA-wang, 2024VITAL-he}, we adopt the case of two views, i.e. view $a$ and view $b$, as an illustrative example for the exposition throughout this paper.
A summary of these notations is provided in Table \ref{tab:notaions}

In this work, we propose the SMART method, as illustrated in Fig. \ref{fig:overview}, to address the above mentioned PVC tasks.
Firstly, we perform representation learning for each view. Specifically, autoencoders are widely used in multi-view learning as an effective technique for compressing raw data into low-dimensional latent representations, facilitating better feature extraction. Typically, view-specific autoencoder can be applied to each view, ensuring that the latent representations capture essential features. More precisely, for each view, an encoder $\mathcal{E}^v$ maps the raw input data $\textbf{X}^v$ into the latent space: $\textbf{Z}^v = \mathcal{E}^v (\textbf{X}^v) \in \mathbb{R}^{N \times d}$, where $v \in \{1, 2, \dots, V\}$, $\textbf{Z}^v =[\textbf{z}^v_1, \textbf{z}^v_2,..., \textbf{z}^v_N] $ and $d$ denotes the shared dimension of the latent embeddings, which remains consistent across views.
Subsequently, a view-specific decoder $\mathcal{D}^v$ is applied to reconstruct data based on these latent features: $\widehat{\textbf{X}}^v = \mathcal{D}^v (\textbf{Z}^v)$. Additionally, the reconstruction loss is introduced to quantify the discrepancy between the original data and their reconstructions for representation learning:
\begin{equation}\label{Eq:loss_rec}
	\mathcal{L}_{rec} = \sum_{v=1}^{V}\left\|\textbf{{X}}^v-\widehat{\textbf{X}}^v\right\|^{2}_{F}.
\end{equation}
The loss can preserve the view-specific information within the latent representations during the reconstruction process.

\subsection{View Distribution Alignment}
For MVC tasks, the key challenge lies in discovering consistent information across all views. However, directly extracting common semantics from heterogeneous data remains challenging, even when learning latent features with identical dimensionality, particularly in PVC tasks, where only a limited number of aligned samples are available.
Given that features of the same instance across different views exhibit distinct heterogeneous characteristics while sharing identical semantic information, the fundamental challenge is how to establish a robust metric that can effectively capture these inherent cross-view semantic relationships and bridge the distributional gap between heterogeneous views by leveraging this metric.
Specifically, since we expect that samples with known alignment relationships exhibit strong intrinsic correlations, we first quantify the cross-view correlation between $\widetilde{\textbf{z}}^a_i$ and $\widetilde{\textbf{z}}^b_j$ within aligned latent embeddings $\widetilde{\textbf{Z}}^v \in \mathbb{R}^{N_a \times d}$ using normalized covariance $\mathbf{\widetilde{\mathcal{C}}}^{ab}_{ij}$, which corresponds to the Pearson's correlation coefficient, a scale-invariant measure. This is defined as: 
\begin{equation}\label{Eq:cov_12}
\begin{split}
    \mathbf{\widetilde{\mathcal{C}}}^{ab}_{ij} &= \frac{\mathrm{cov}(\widetilde{\textbf{z}}^{a}_i, \widetilde{\textbf{z}}^{b}_j)} {\sigma (\widetilde{\textbf{z}}^a_i) \sigma (\widetilde{\textbf{z}}^b_j)}
    \\ &= \frac{1}{d-1} \left( \frac{\widetilde{\textbf{z}}^{a}_i - \mu (\widetilde{\textbf{z}}^a_i)}{\sigma (\widetilde{\textbf{z}}^a_i)} \right) \left( \frac{\widetilde{\textbf{z}}^{b}_j - \mu (\widetilde{\textbf{z}}^b_j)}{\sigma (\widetilde{\textbf{z}}^b_j)} \right) ^ {\top},
\end{split}
\end{equation}
where $\mathbf{\widetilde{\mathcal{C}}}^{ab} \in \mathbb{R}^{N_a \times N_a}$ and $\mathrm{cov}(\widetilde{\textbf{z}}^{a}_i, \textbf{z}^{b}_j)$ is the cross-view covariance between latent representation distributions $\widetilde{\textbf{z}}^{a}_i$ and $\widetilde{\textbf{z}}^{b}_j$, $\mu(\cdot)$ and $\sigma(\cdot)$ represent the mean value and standard deviation, respectively.

Cross-view samples with known alignment relationships exhibit strong intrinsic correlations, such that the cross-view covariance approximates an identity vector. This indicates that each aligned sample pair contains identical semantic information. To achieve this, we introduce the following cross-view feature alignment loss:
\begin{equation}\label{Eq:loss_diag}
    \mathcal{L}_{cfa} = \frac{1}{N_a} \left\| \mathrm{diag}(\mathbf{\widetilde{\mathcal{C}}}^{ab}) - \mathbf{1} \right\|^{2}_{2},
\end{equation}
where $\mathrm{diag(\cdot)}$ is a row vector containing the diagonal elements, and $\textbf{1} \in \mathbb{R}^{N_a}$ is a row vector with all elements equal to 1.
This loss function enforces linear consistency between feature variations across views for aligned instances, promoting view-invariant representation learning and semantic consistency.
\begin{figure*}[t]
    \centering
    \includegraphics[width=1.0\linewidth]{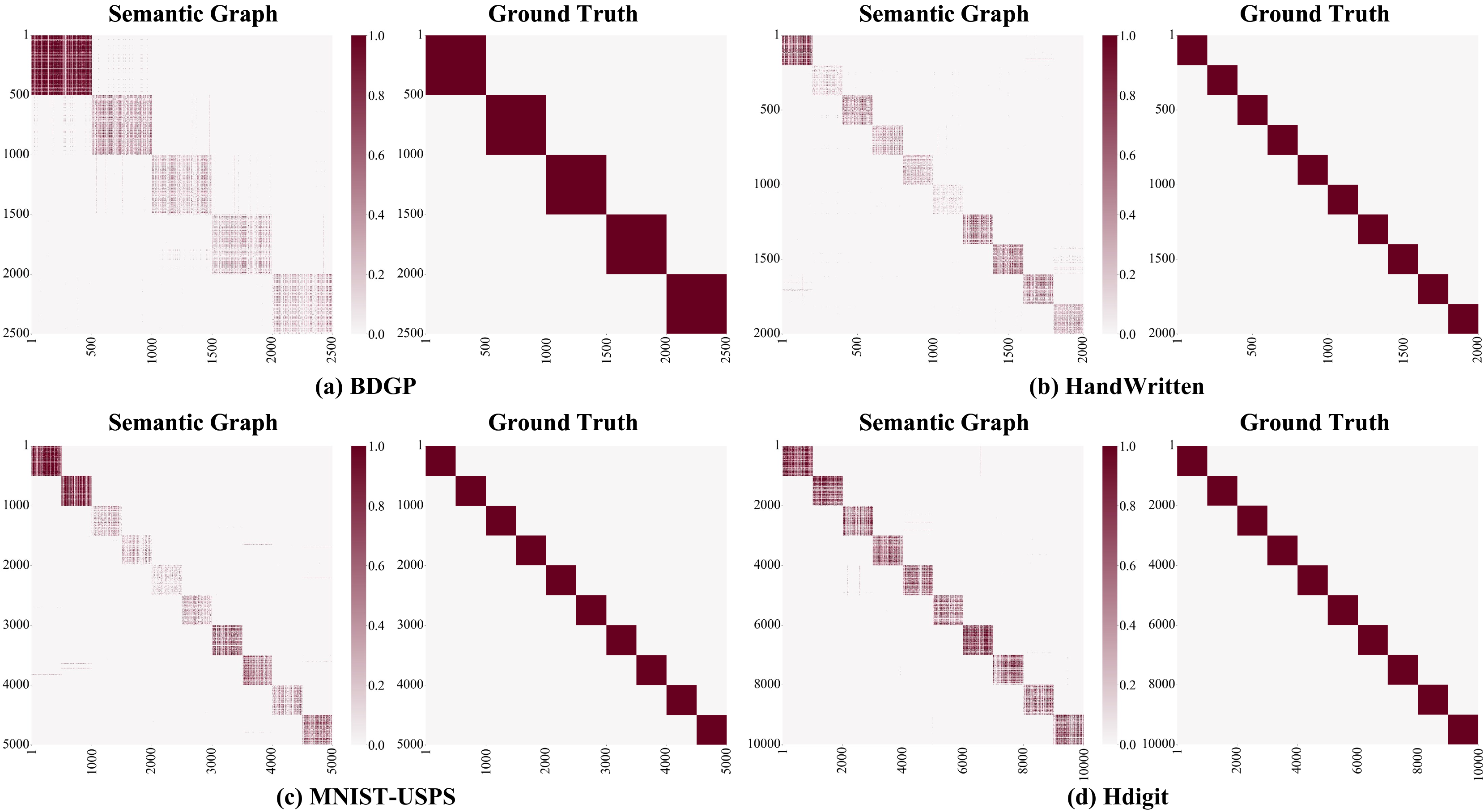}
    \caption{Visualization of the rearranged semantic guidance graphs $\mathbf{\Omega}$ in Eq. (\ref{Eq:adj}) alongside the ground-truth category matrices on the complete (a) BDGP, (b) HandWritten, (c) MNIST-USPS, and (d) Hdigit datasets. Each point in the semantic graphs represents a neighbor pair, and darker color indicate larger edge weight (within the range of $[0, 1]$ and stronger correlation. 
    Evidently, the learned semantic graphs exhibit strong consistency with the true class structure, with high correlation values concentrated in regions corresponding to ground-truth categories.
    }
    \label{fig:adj}
\end{figure*}
In addition, if the embeddings $\widetilde{\textbf{z}}^a_i$ and $\widetilde{\textbf{z}}^a_j$ share similar semantic characteristics, then their counterparts $\widetilde{\textbf{z}}^b_i$ and $\widetilde{\textbf{z}}^b_j$ should preserve the same similarity structure. To capture such generalized, cluster-level similarity structures and further reduce cross-view discrepancies, we adopt a covariance matching metric.
Specifically, let $\mathbf{\widetilde{\mathcal{C}}}^{v} \in \mathbb{R}^{N_a \times N_a}$ denotes the intra-view normalized covariance matrix computed from the aligned latent embeddings  $\widetilde{\textbf{Z}}^v$of the $v$-th view, defined as:
\begin{equation}\label{Eq:cov_v}
\begin{split}
    \mathbf{\widetilde{\mathcal{C}}}^{v}_{ij} &= \frac{\mathrm{cov}(\widetilde{\textbf{z}}^{v}_i, \widetilde{\textbf{z}}^{v}_j)} {\sigma (\widetilde{\textbf{z}}^v_i) \sigma (\widetilde{\textbf{z}}^v_j)}.
\end{split}
\end{equation}
Thus, this covariance matching alignment can be expressed as the approximate equality of corresponding matrix elements, i.e., $\mathbf{\widetilde{\mathcal{C}}}^{a}_{ij} \approx \mathbf{\widetilde{\mathcal{C}}}^{b}_{ij}$ for all aligned instance pairs $(i,j)$, which could be formally defined as:
\begin{equation}\label{Eq:loss_CC}
    \mathcal{L}_{cma} = \frac{1}{N_a} \left\| \mathbf{\widetilde{\mathcal{C}}}^{a} - \mathbf{\widetilde{\mathcal{C}}}^{b} \right\|^{2}_{F}. 
\end{equation}

Subsequently, the view distribution alignment can be achieved through the joint optimization of $\mathcal{L}_{cma}$ and $\mathcal{L}_{cfa}$, expressed as:
\begin{equation}\label{Eq:loss_cov}
\begin{split}
    \mathcal{L}_{vda} &= \mathcal{L}_{cfa} + \mathcal{L}_{cma} \\ 
    &= \frac{1}{N_a} \left( \left\| \mathrm{diag}(\mathbf{\widetilde{\mathcal{C}}}^{ab}) - \mathbf{1} \right\|^{2}_{2} + \left\| \mathbf{\widetilde{\mathcal{C}}}^{a} - \mathbf{\widetilde{\mathcal{C}}}^{b} \right\|^{2}_{F} \right).
\end{split}
\end{equation}
The two alignment losses jointly consider cross-view covariance and the minimization of discrepancies between view-specific covariance matrices. Compared to instance-level alignment methods i.e. MMD, covariance alignment can be regarded as a second-order statistic, capable of capturing richer structural relationships between instances. By optimizing such metrics, the method effectively reduces the distributional gap between views, thereby enhancing cross-view consistency. 

\subsection{Semantic Matching Contrastive Learning}
Although cross-view semantic correlation alignment, as defined in Eq. (\ref{Eq:loss_cov}), reduces distribution mismatch across different views using aligned data, it fails to fully capture consistency information in unaligned data, thus limiting its ability to enhance clustering performance. To address this, we further propose leveraging adaptive graph contrastive learning \cite{2025NeuCGC-peng}, based on a learned cross-view semantic guidance graph, to exploit consistency information in both aligned and unaligned samples. Specifically, we utilize a shared MLP projector to map latent embeddings into a unified high-level feature space, followed by $l_2$-normalization to obtain the representations:
\begin{equation}\label{Eq:p_high_level}
    \textbf{H}^{v} = \mathcal{P}(\textbf{Z}^{v}) ,
\end{equation}
where $\textbf{H}^v \in \mathbb{R}^{N \times d}$ is the high-level features of $v$-th view obtained by the shared projector $\mathcal{P}$. This transformation filters out irrelevant details and reduces redundancy, enhancing the model's ability to capture abstract semantic consistency across all views. 

Building on reliable cross-view semantic correlations from aligned data in Eq. (\ref{Eq:cov_12}), we then refine a cross-view semantic guidance graph $\mathbf{\Omega}^{ab} = [\omega^{ab}_1, \omega^{ab}_2, ..., \omega^{ab}_N] \in \mathbb{R}^{N \times N}$ from the pairwise normalized covariance matrix $\mathbf{\mathcal{C}}^{ab} \in \mathbb{R}^{N \times N}$ over all samples. Given $i,j = 1, 2,...,N$,
\begin{equation}\label{Eq:adj}
    \mathbf{\omega}^{ab}_{ij} = \begin{cases}
    1, & i = j ~ and ~ \textbf{x}^a_i \sim \textbf{x}^b_j, \\
    \mathbf{\mathcal{C}}_{ij}^{ab},~ & \mathbf{\mathcal{C}}_{ij}^{ab} > \mathcal{T}^{ab}, \\
    0, & otherwise,
    \end{cases}
\end{equation}
where $0 \le \mathbf{\omega}^{ab}_{ij} \le 1$, $\textbf{x}^a_i\sim\textbf{x}^b_i$ represents that $\textbf{x}^a_i$ and $\textbf{x}^b_i$ are aligned, and $\mathcal{T} \in (0, 1)$ is a predefined threshold that determines the minimum correlation level required for two samples to be considered semantically similar graph neighbors, as visualized in Fig. \ref{fig:adj}. Intuitively, we use the aligned samples as reference distribution, selecting as reliable neighbors those portions exceeding the mean covariance minus one standard deviation. It could be formally defined as:
\begin{equation}\label{eq1}
    \mathcal{T}^{ab} = \mathrm{max}\left(0, \mu(\mathrm{diag}(\mathbf{\widetilde{\mathcal{C}}}^{ab})) - \sigma (\mathrm{diag}(\mathbf{\widetilde{\mathcal{C}}}^{ab})) \right).
\end{equation}
We then leverage the topological structure and edge weights of the semantic guidance graph $\mathbf{\Omega}^{ab}$ to guide contrastive learning in exploiting cross-view consistency for all samples. 

Specifically, we define the representations of aligned instances from different views as positive pairs. Simultaneously, neighboring instances within the semantic graph are treated as semantic pairs, which are assigned weighted positive pair values ranging from 0 to 1 for mitigating interference from noisy neighbors. All other instance pairs are considered negative. This enables us to formulate the semantic matching contrastive loss between view $a$ and view $b$ as follows:
\begin{equation}\label{Eq:cl}
\begin{split}
    \mathcal{L}_{smc} &= \frac{1}{N} \sum_{i=1}^{N} - \mathrm{log} \frac{\sum_{k \in \mathcal{N}^{ab}_i\cup i} \mathbf{\omega}^{ab}_{ik} \mathrm{exp}(s(\textbf{h}_i^a, \textbf{h}_k^b) / \tau) }{\sum_{j=1}^N \mathrm{exp}(s(\textbf{h}_i^a, \textbf{h}_j^b) / \tau)} \\
&= \frac{1}{N} \sum_{i=1}^{N} \\ - \mathrm{log} &\frac{\mathds{1} \mathrm{exp}(s(\textbf{h}_i^a, \textbf{h}_i^b) / \tau) + \sum_{k \in \mathcal{N}^{ab}_i} \mathbf{\omega}^{ab}_{ik} \mathrm{exp}(s(\textbf{h}_i^a, \textbf{h}_k^b) / \tau) }{\sum_{j=1}^N \mathrm{exp}(s(\textbf{h}_i^a, \textbf{h}_j^b) / \tau)}, 
\end{split}
\end{equation}
where $\mathcal{N}^{ab}_i$ is the neighborhood set of $i$-th sample within $\mathbf{\Omega}^{ab}$, $\mathds{1}$ represents the indicator function $\mathds{1}(\textbf{x}^a_i \sim \textbf{x}^b_i)$ which means $\textbf{x}^a_i$ and $\textbf{x}^b_i$ are aligned with each other, $s(\cdot)$ is the cosine similarity, 
namely $s(\mathbf{h}_i^a, \mathbf{h}_j^b)=\frac{\mathbf{h}^{a}_i (\mathbf{h}^{b}_j)^{\top}}{||\mathbf{h}^{a}_i||_2||\mathbf{h}^{b}_j||_2}$, 
and $\tau$ denotes the temperature parameter.
\begin{algorithm}[!t]
\caption{Optimization of SMART}
\label{algorithm}
\textbf{Input}: Partially view-aligned data $\{\textbf{X}^v\}_{v=1}^V$; iteration number $I$; batch size $B$; embeddings dimension $d$; hyper-parameters $\lambda_1$ and $\lambda_2$.
\begin{algorithmic}[1] 
\FOR{epoch = 1 to $I$}
\STATE Obtain $\textbf{Z}^v$ by encoding $\textbf{X}^v$ with $\mathcal{E}^v$.
\STATE Restore $\textbf{X}^v$ by $\widehat{\mathbf{X}}^v$ from $\textbf{Z}^v$ using $\mathcal{D}^v$, then calculate $\mathcal{L}_{rec}$ with Eq. (\ref{Eq:loss_rec}).
\STATE Calculate $\mathbf{\widetilde{\mathcal{C}}}^{ab}$ and $\mathbf{\widetilde{\mathcal{C}}}^{v}$ of aligned samples with Eqs. (\ref{Eq:cov_12}, \ref{Eq:cov_v}), respectively. Then calculate $\mathcal{L}_{vda}$ via Eq. (\ref{Eq:loss_cov}).
\STATE Learn $\mathbf{\Omega}^{ab}$ over all samples with $\mathcal{T}$ via Eq. (\ref{Eq:adj}).
\STATE Project $\textbf{Z}^v$ to $\textbf{H}^v$, then calculate $\mathcal{L}_{smc}$ via Eq. (\ref{Eq:cl}).
\STATE Optimize the entire networks by jointly minimizing $\mathcal{L}_{rec}$, $\mathcal{L}_{vda}$ and $\mathcal{L}_{smc}$.
\ENDFOR
\STATE Perform K-means algorithm on the fused representations $\textbf{H}$ via Eq. (\ref{Eq:fusion}) to generate cluster assignments $\textbf{c}$.
\end{algorithmic}
\textbf{Output}: {Clustering results $\textbf{c}$.}
\end{algorithm}

\begin{table*}[t]
\caption{Clustering performance (\%) on partially aligned (50\%) multi-view datasets (mean $\pm$ std. over 5 runs). Best and second-best results highlighted in bold and underline, respectively.}
\centering
\scalebox{0.83}{
\fontsize{9.5}{11}\selectfont 
\setlength{\tabcolsep}{3.5pt}
\begin{tabular}{@{}rr|ccc|ccc|ccc|ccc@{}}
\toprule[1.25pt]
\multicolumn{2}{c|}{\multirow{2}{*}{\textbf{Methods}}} & \multicolumn{3}{c|}{\textbf{HandWritten}} & \multicolumn{3}{c|}{\textbf{BDGP}} & \multicolumn{3}{c|}{\textbf{WIKI}} & \multicolumn{3}{c}{\textbf{MNIST-USPS}} \\

& &\textbf{ACC} & \textbf{NMI} & \textbf{ARI} & \textbf{ACC} & \textbf{NMI} & \textbf{ARI} & \textbf{ACC} & \textbf{NMI} & \textbf{ARI} & \textbf{ACC} & \textbf{NMI} & \textbf{ARI} \\
\midrule
(CVPR'22)& MFLVC & 34.77{\tiny$\pm$1.60} & 24.37{\tiny$\pm$0.95} & 15.83{\tiny$\pm$0.64} & 73.47{\tiny$\pm$7.71} & 60.55{\tiny$\pm$6.91} & 55.68{\tiny$\pm$9.68} & 24.79{\tiny$\pm$2.73} & 9.94{\tiny$\pm$1.45} & 4.66{\tiny$\pm$1.50} & 60.02{\tiny$\pm$1.29} & 37.00{\tiny$\pm$1.30} & 32.99{\tiny$\pm$1.04}\\

(MM'23) & DealMVC & 37.03{\tiny$\pm$1.11} & 25.61{\tiny$\pm$0.65} & 17.25{\tiny$\pm$0.30} & 60.74{\tiny$\pm$4.37} & 47.40{\tiny$\pm$2.72} & 37.65{\tiny$\pm$3.00} & 32.63{\tiny$\pm$0.51} & 14.49{\tiny$\pm$0.14} & 10.39{\tiny$\pm$0.29} & 60.48{\tiny$\pm$1.76} & 35.65{\tiny$\pm$1.00} & 32.53{\tiny$\pm$1.28} \\

(CVPR'23) & GCFAgg & 39.04{\tiny$\pm$0.30} & 26.55{\tiny$\pm$0.70} & 17.05{\tiny$\pm$0.35} & 46.82{\tiny$\pm$6.23} & 31.01{\tiny$\pm$5.40} & 21.87{\tiny$\pm$6.24} & 29.18{\tiny$\pm$1.67} & 11.15{\tiny$\pm$0.68} & 7.09{\tiny$\pm$0.54} & 56.84{\tiny$\pm$0.44} & 30.60{\tiny$\pm$0.49} & 27.36{\tiny$\pm$0.54} \\

(AAAI'24) & ICMVC & 48.28{\tiny$\pm$3.36} & 36.34{\tiny$\pm$1.04} & 25.63{\tiny$\pm$1.67} & 63.90{\tiny$\pm$2.24} & 40.44{\tiny$\pm$3.94} & 37.99{\tiny$\pm$4.31} & 23.96{\tiny$\pm$0.94} & 8.34{\tiny$\pm$0.84} & 5.44{\tiny$\pm$0.68} & 70.32{\tiny$\pm$0.62} & 46.36{\tiny$\pm$0.83} & 45.11{\tiny$\pm$0.94} \\

(AAAI'24) & SURER & 72.76{\tiny$\pm$5.54} & 63.66{\tiny$\pm$5.57} & 49.90{\tiny$\pm$9.02} & 42.50{\tiny$\pm$5.82} & 17.26{\tiny$\pm$7.50} & 12.22{\tiny$\pm$5.01} & 32.77{\tiny$\pm$0.80} & 15.11{\tiny$\pm$0.45} & 10.56{\tiny$\pm$0.56} & 32.31{\tiny$\pm$3.91} & 23.10{\tiny$\pm$7.42} & 12.18{\tiny$\pm$3.36} \\

(CVPR'24) & MVCAN &40.37{\tiny$\pm$1.88} & 29.27{\tiny$\pm$0.72} & 19.83{\tiny$\pm$0.60} & 71.20{\tiny$\pm$6.59} & 54.30{\tiny$\pm$7.36} & 50.25{\tiny$\pm$6.98} & 31.95{\tiny$\pm$0.24} & 15.05{\tiny$\pm$0.08} & 10.26{\tiny$\pm$0.08} & 61.50{\tiny$\pm$4.61} & 51.51{\tiny$\pm$5.41} & 43.53{\tiny$\pm$6.96}\\
\midrule
(NIPS'20) & PVC & 84.16{\tiny$\pm$0.51} & 78.04{\tiny$\pm$0.38} & 74.34{\tiny$\pm$0.52} & 85.53{\tiny$\pm$0.03} & 64.77{\tiny$\pm$0.06} & 67.18{\tiny$\pm$0.06} & 28.19{\tiny$\pm$0.11} & 18.08{\tiny$\pm$0.17} & 10.55{\tiny$\pm$0.13} & 75.85{\tiny$\pm$0.41} & 71.05{\tiny$\pm$0.31} & 61.34{\tiny$\pm$0.39}\\

(CVPR'21) & MvCLN & 63.66{\tiny$\pm$0.55} & 58.98{\tiny$\pm$0.55} & 47.77{\tiny$\pm$0.52} & 80.61{\tiny$\pm$1.03} & 58.26{\tiny$\pm$2.05} & 58.49{\tiny$\pm$1.97} & 35.52{\tiny$\pm$0.48} & 17.22{\tiny$\pm$0.43} & 12.82{\tiny$\pm$0.55} & 85.60{\tiny$\pm$0.49} & 72.60{\tiny$\pm$0.71} & 71.23{\tiny$\pm$0.82}\\

(TPAMI'23) & SURE & 78.09{\tiny$\pm$0.24} & 73.74{\tiny$\pm$0.61} & 62.47{\tiny$\pm$0.39} & 82.46{\tiny$\pm$0.76} & 66.38{\tiny$\pm$0.97} & 59.52{\tiny$\pm$1.34} & 34.86{\tiny$\pm$0.07} & 16.22{\tiny$\pm$0.29} & 12.27{\tiny$\pm$0.27} & 93.35{\tiny$\pm$0.29} & 85.40{\tiny$\pm$0.57} & 85.98{\tiny$\pm$0.60} \\

(MM'24) & CGDA 
&83.37{\tiny$\pm$3.58} &77.91{\tiny$\pm$3.88} &71.88{\tiny$\pm$4.56} &90.56{\tiny$\pm$0.15} &77.53{\tiny$\pm$0.68} &76.83{\tiny$\pm$0.97} &40.40{\tiny$\pm$1.42} &34.52{\tiny$\pm$1.04} &26.83{\tiny$\pm$0.33} &80.11{\tiny$\pm$4.51} &80.21{\tiny$\pm$1.75} &72.07{\tiny$\pm$3.97}	\\

(MM'24) & VITAL 
&89.61{\tiny$\pm$0.61} &80.16{\tiny$\pm$0.95} &78.31{\tiny$\pm$1.05} &82.91{\tiny$\pm$6.41} &70.34{\tiny$\pm$6.07} &66.51{\tiny$\pm$7.40} &36.44{\tiny$\pm$0.17} &20.70{\tiny$\pm$0.65} &15.24{\tiny$\pm$0.44} &\underline{94.28{\tiny$\pm$0.45}} &86.16{\tiny$\pm$0.83} &87.75{\tiny$\pm$0.92} \\

(TCSVT'24) &CGCN  
&71.18{\tiny$\pm$2.22} &72.55{\tiny$\pm$1.30} &62.24{\tiny$\pm$2.28} &88.06{\tiny$\pm$0.27} &70.14{\tiny$\pm$0.51} &72.86{\tiny$\pm$0.57} &35.71{\tiny$\pm$0.34} &19.01{\tiny$\pm$0.23} &13.85{\tiny$\pm$0.40} &94.09{\tiny$\pm$0.03} &\underline{88.96{\tiny$\pm$0.03}} &\underline{89.59{\tiny$\pm$0.06}} \\

(TCSVT'25) &TCLPVC  
&\underline{92.68{\tiny$\pm$1.27}} &\underline{86.38{\tiny$\pm$1.24}} &\underline{85.62{\tiny$\pm$1.42}} &\underline{91.78{\tiny$\pm$0.24}} &\underline{77.94{\tiny$\pm$0.78}} &\underline{80.59{\tiny$\pm$0.56}} &\underline{41.00{\tiny$\pm$2.79}} &\underline{35.72{\tiny$\pm$2.38}} &\underline{27.02{\tiny$\pm$3.71}} &93.70{\tiny$\pm$0.26} &86.12{\tiny$\pm$0.42} &86.51{\tiny$\pm$0.53} \\

\rowcolor{green!10} (\textbf{Ours}) & \textbf{SMART} & \textbf{93.27{\tiny$\pm$0.67}} & \textbf{86.62{\tiny$\pm$0.96}} & \textbf{85.62{\tiny$\pm$1.34}} & \textbf{95.58{\tiny$\pm$0.52}} & \textbf{89.44{\tiny$\pm$1.48}} & \textbf{89.47{\tiny$\pm$1.22}} & \textbf{46.16{\tiny$\pm$0.91}} & \textbf{40.34{\tiny$\pm$1.27}} & \textbf{29.57{\tiny$\pm$1.09}} & \textbf{95.77{\tiny$\pm$0.72}} & \textbf{89.99{\tiny$\pm$1.02}} & \textbf{90.91{\tiny$\pm$1.44}}  \\ 
\midrule[0.85pt]
\multicolumn{2}{c|}{\multirow{2}{*}{\textbf{Methods}}} & \multicolumn{3}{c|}{\textbf{NUS-WIDE}} & \multicolumn{3}{c|}{\textbf{Hdigit}} & \multicolumn{3}{c|}{\textbf{Deep Animal}} & \multicolumn{3}{c}{\textbf{Reuters}} \\

& &\textbf{ACC} & \textbf{NMI} & \textbf{ARI} & \textbf{ACC} & \textbf{NMI} & \textbf{ARI} & \textbf{ACC} & \textbf{NMI} & \textbf{ARI} & \textbf{ACC} & \textbf{NMI} & \textbf{ARI} \\
\midrule
(CVPR'22) &MFLVC 
&16.45{\tiny$\pm$0.34} &3.86{\tiny$\pm$0.54} &2.35{\tiny$\pm$0.18} &78.48{\tiny$\pm$1.24} &59.98{\tiny$\pm$1.68} &58.55{\tiny$\pm$2.03} &12.01{\tiny$\pm$1.08} &20.87{\tiny$\pm$1.15} &6.39{\tiny$\pm$0.42} &38.04{\tiny$\pm$1.57} &17.13{\tiny$\pm$1.42} &12.83{\tiny$\pm$0.84} \\

(MM'23) & DealMVC 
&26.98{\tiny$\pm$1.64} &9.21{\tiny$\pm$0.73} &6.02{\tiny$\pm$0.63} &75.53{\tiny$\pm$5.89} &62.50{\tiny$\pm$3.23} &59.61{\tiny$\pm$4.95} &29.09{\tiny$\pm$1.35} &38.05{\tiny$\pm$0.70} &16.64{\tiny$\pm$0.61} &45.06{\tiny$\pm$2.47} &15.05{\tiny$\pm$1.84} &15.69{\tiny$\pm$1.25} \\

(CVPR'23) & GCFAgg 
&21.84{\tiny$\pm$3.86} &9.57{\tiny$\pm$3.71} &5.32{\tiny$\pm$2.34} &79.68{\tiny$\pm$2.04} &62.12{\tiny$\pm$3.23} &60.57{\tiny$\pm$3.53} &18.96{\tiny$\pm$1.78} &31.27{\tiny$\pm$1.68} &10.40{\tiny$\pm$1.47} &36.42{\tiny$\pm$1.71} &15.61{\tiny$\pm$1.09} &12.05{\tiny$\pm$0.94} \\

(AAAI'24) & ICMVC 
&42.45{\tiny$\pm$0.57} &20.71{\tiny$\pm$0.49} &16.98{\tiny$\pm$0.49} &78.99{\tiny$\pm$0.81} &58.81{\tiny$\pm$1.28} &59.00{\tiny$\pm$1.39} &43.51{\tiny$\pm$2.01} &53.73{\tiny$\pm$0.84} &31.67{\tiny$\pm$1.75} &39.24{\tiny$\pm$1.68} &15.82{\tiny$\pm$1.52} &12.79{\tiny$\pm$0.89} \\

(AAAI'24) & SURER 
&23.07{\tiny$\pm$2.72} &13.58{\tiny$\pm$2.88} &5.63{\tiny$\pm$2.25} &29.12{\tiny$\pm$2.49} &17.85{\tiny$\pm$3.43} &8.03{\tiny$\pm$0.98} &13.41{\tiny$\pm$1.22} &19.11{\tiny$\pm$1.23} &3.92{\tiny$\pm$0.90} &37.85{\tiny$\pm$3.72} &10.11{\tiny$\pm$1.58} &9.25{\tiny$\pm$3.48} \\

(CVPR'24) & MVCAN 
&26.27{\tiny$\pm$0.78} &8.97{\tiny$\pm$0.53} &5.48{\tiny$\pm$0.47} &39.65{\tiny$\pm$2.94} &18.71{\tiny$\pm$1.24} &14.36{\tiny$\pm$1.44} &40.10{\tiny$\pm$2.55} &57.39{\tiny$\pm$2.46} &34.16{\tiny$\pm$3.03} &34.52{\tiny$\pm$1.63} &10.47{\tiny$\pm$0.73} &10.02{\tiny$\pm$0.84} \\

\midrule

(NIPS'20) & PVC 
&44.01{\tiny$\pm$0.51} &39.43{\tiny$\pm$0.45} &32.78{\tiny$\pm$0.24} &81.92{\tiny$\pm$0.03} &81.22{\tiny$\pm$0.04} &80.26{\tiny$\pm$0.05} &3.83{\tiny$\pm$0.00} &0.00{\tiny$\pm$0.00} &0.00{\tiny$\pm$0.00} &41.95{\tiny$\pm$0.37} &19.80{\tiny$\pm$0.54} &17.12{\tiny$\pm$0.48} \\

(CVPR'21) & MvCLN 
&62.68{\tiny$\pm$0.35} &44.73{\tiny$\pm$0.52} &40.52{\tiny$\pm$0.58} &94.94{\tiny$\pm$0.42} &88.15{\tiny$\pm$0.89} &89.12{\tiny$\pm$0.89} &36.72{\tiny$\pm$0.30} &55.41{\tiny$\pm$0.36} &31.60{\tiny$\pm$0.33} &50.27{\tiny$\pm$1.96} &29.21{\tiny$\pm$1.82} &24.95{\tiny$\pm$2.07} \\

(TPAMI'23) & SURE 
&59.37{\tiny$\pm$0.33} &41.39{\tiny$\pm$0.43} &37.83{\tiny$\pm$0.36} &93.04{\tiny$\pm$0.33} &84.14{\tiny$\pm$0.56} &85.21{\tiny$\pm$0.65} &28.44{\tiny$\pm$0.17} &40.16{\tiny$\pm$0.31} &19.11{\tiny$\pm$0.36} &50.13{\tiny$\pm$2.03} &28.81{\tiny$\pm$1.97} &24.73{\tiny$\pm$2.22} \\

(MM'24) & CGDA &42.96{\tiny$\pm$0.10} &34.90{\tiny$\pm$0.14} &28.05{\tiny$\pm$0.06} &74.49{\tiny$\pm$3.01} &81.95{\tiny$\pm$4.11} &70.62{\tiny$\pm$4.71} &21.68{\tiny$\pm$0.65} &35.24{\tiny$\pm$0.24} &15.17{\tiny$\pm$0.63} & &OOM &  \\

(MM'24) & VITAL 
&\underline{62.85{\tiny$\pm$1.00}} &\underline{47.75{\tiny$\pm$1.16}} &\underline{42.08{\tiny$\pm$1.41}} &96.44{\tiny$\pm$0.17} &90.60{\tiny$\pm$0.40} &92.25{\tiny$\pm$0.36} &44.79{\tiny$\pm$0.48} &49.68{\tiny$\pm$0.59} &30.11{\tiny$\pm$0.49} &50.56{\tiny$\pm$5.04} &33.80{\tiny$\pm$2.16} &22.97{\tiny$\pm$3.32} \\

(TCSVT'24) &CGCN  
&56.45{\tiny$\pm$1.50} &39.47{\tiny$\pm$0.31} &34.01{\tiny$\pm$0.90} &\underline{96.55{\tiny$\pm$0.10}} &\underline{91.34{\tiny$\pm$0.22}} &\underline{92.52{\tiny$\pm$0.22}} &38.76{\tiny$\pm$1.02} &46.53{\tiny$\pm$0.42} &24.96{\tiny$\pm$0.92} &50.84{\tiny$\pm$0.82} &30.91{\tiny$\pm$0.93} &25.36{\tiny$\pm$0.66} \\

(TCSVT'25) &TCLPVC  
&44.47{\tiny$\pm$3.04} &33.28{\tiny$\pm$1.42} &25.53{\tiny$\pm$1.76} &95.54{\tiny$\pm$0.25} &89.52{\tiny$\pm$0.41} &90.37{\tiny$\pm$0.52} &\underline{55.12{\tiny$\pm$1.27}} &\underline{68.88{\tiny$\pm$0.39}} &\underline{46.50{\tiny$\pm$1.32}} &\underline{54.57{\tiny$\pm$2.17}} &\textbf{35.73{\tiny$\pm$2.88}} &\underline{29.49{\tiny$\pm$2.34}} \\

 \rowcolor{green!10} (\textbf{Ours}) & \textbf{SMART} &\textbf{64.84{\tiny$\pm$0.43}} &\textbf{49.88{\tiny$\pm$0.47}} &\textbf{44.95{\tiny$\pm$0.47}} &\textbf{98.00{\tiny$\pm$0.24}} &\textbf{94.73{\tiny$\pm$0.44}} &\textbf{95.61{\tiny$\pm$0.51}} &\textbf{63.07{\tiny$\pm$0.98}} &\textbf{68.96{\tiny$\pm$0.72}} &\textbf{52.32{\tiny$\pm$1.28}} &\textbf{58.40{\tiny$\pm$1.74}} &\underline{35.60{\tiny$\pm$3.07}} &\textbf{35.06{\tiny$\pm$1.42}} \\ 
\bottomrule[1.25pt]
\end{tabular}
}
\label{tab:methods_comparison_50}
\end{table*}
As a result, we are able to maximize feature consistency within positive pairs, partially maximize consistency within semantic pairs, and minimize consistency within negative pairs. This approach enhances the discriminative capability of the model, irrespective of whether the samples are aligned or unaligned. \textit{\textbf{Notably, in contrast to existing PVC methods, our approach integrates both the semantic guidance graph and the proposed contrastive loss applied to the entire dataset (including both aligned and unaligned samples), without overly relying on learned view correspondences among unaligned samples, which may be distorted by inaccurate view matching.}} This allows our method to produce more comprehensive feature representations while simultaneously improving the robustness of feature consistency learning.

\noindent \textbf{Semantic Matching Feature Fusion.} 
Finally, leveraging the learned semantic guidance graph $\mathbf{\Omega}^{ab}$, we perform semantic matching feature fusion and apply the K-means algorithm to the fused representations for clustering. Specifically, by concatenating a baseline view $\textbf{h}^a_i$ and the matched representation $\widehat{\textbf{h}}^b_i = \sum_{k \in \mathcal{N}_i^{ab}} \mathbf{\omega}^{ab}_{ik} \textbf{h}^b_k$, we obtain the final fused representations $\textbf{H} = [\textbf{h}_1, \textbf{h}_2, ..., \textbf{h}_N]$ from all views as follows:
\begin{equation}\label{Eq:fusion}
    \textbf{h}_i = \mathrm{cat}(\textbf{h}^1_i, \mathrm{norm}(\widehat{\textbf{h}}^2_i),..., \mathrm{norm}(\widehat{\textbf{h}}^V_i)),
\end{equation}
where $\mathrm{cat}(\cdot)$ is the concatenation operator, and $\mathrm{norm}(\cdot)$ denotes the $l_2$-normalization. 
Here, under the guidance of $\mathbf{\Omega}^{ab}$, the fused representations $\textbf{H}$ effectively aggregate common features from the most reliable and semantically coherent instances without requiring additional distance computation and sample-to-sample view realignment adopted by prior works \cite{2020PVC-huang, 2021MvCLN-yang}. This results in more discriminative and cluster-representative features, ultimately leading to higher-quality cluster assignments.

\begin{table}[!t]
\centering
\caption{Summary of Dataset Statistics.}
    \scalebox{1.05}{
    \renewcommand{\arraystretch}{1.1} 
    \begin{tabular}{*{5}{c}}
        \toprule[1.25pt]
        Dataset &\#Sample &\#Class &\#View  &Type \\
        \midrule
        HandWritten \cite{2023HandWritten-wen} &2000 &10 &6 &Image \\
        BDGP \cite{2012BDGP-cai} &2500 &5 &2 &Multimodal \\
        WIKI \cite{2010WIKI-Rasiwasia} &2866 &10 &2 &Multimodal \\
        MNIST-USPS \cite{2019MNISTUSPS-peng} &5000 &10 &2 &Image \\
        NUS-WIDE \cite{2009NUSWIDE-chua} &9000 &10 &2 &Multimodal \\
        Hdigit \cite{2022Hdigit-chen} &10000 &10 &2 &Image \\
        Deep Animal \cite{2013DeepAnimal-Lampert} &10158 &50 &7 &Multimodal \\
        Reuters \cite{2009Reuters-Amini} &18758 &6 &5 &Text \\
        \bottomrule[1.25pt]
    \end{tabular}
    }
    \label{tab:dataset_info}
\end{table}
\begin{table*}[t]
\caption{Clustering performance (\%) on fully aligned (100\%) multi-view datasets (mean $\pm$ std. over 5 runs). Best and second-best results highlighted in bold and underline, respectively.}
\centering
\scalebox{0.83}{
\fontsize{9.5}{11}\selectfont 
\setlength{\tabcolsep}{3.5pt}
\begin{tabular}{rr|ccc|ccc|ccc|ccc@{}}
\toprule[1.25pt]
\multicolumn{2}{c|}{\multirow{2}{*}{\textbf{Methods}}} & \multicolumn{3}{c|}{\textbf{HandWritten}} & \multicolumn{3}{c|}{\textbf{BDGP}} & \multicolumn{3}{c|}{\textbf{WIKI}} & \multicolumn{3}{c}{\textbf{MNIST-USPS}} \\
& &\textbf{ACC} & \textbf{NMI} & \textbf{ARI} & \textbf{ACC} & \textbf{NMI} & \textbf{ARI} & \textbf{ACC} & \textbf{NMI} & \textbf{ARI} & \textbf{ACC} & \textbf{NMI} & \textbf{ARI} \\
\midrule
(CVPR'22) &MFLVC 
&54.55{\tiny$\pm$2.99} &50.99{\tiny$\pm$2.21} &36.02{\tiny$\pm$2.70} &93.28{\tiny$\pm$4.36} &88.57{\tiny$\pm$5.52} & 85.12{\tiny$\pm$7.21} &40.10{\tiny$\pm$3.27} & 33.83{\tiny$\pm$3.16} &24.27{\tiny$\pm$3.08} & 99.54{\tiny$\pm$0.14} &98.69{\tiny$\pm$0.35} & 99.01{\tiny$\pm$0.29} \\

(MM'23) &DealMVC 
&60.39{\tiny$\pm$5.28} &59.41{\tiny$\pm$1.77} &45.99{\tiny$\pm$3.01} &80.39{\tiny$\pm$4.49} &74.16{\tiny$\pm$5.07} &62.94{\tiny$\pm$8.09} &56.22{\tiny$\pm$2.19} &49.36{\tiny$\pm$1.30} &40.30{\tiny$\pm$1.41} &93.34{\tiny$\pm$5.59} &95.76{\tiny$\pm$4.57} &92.95{\tiny$\pm$6.44} \\

(CVPR'23) &GCFAgg 
&64.96{\tiny$\pm$7.51} &59.88{\tiny$\pm$3.93} &47.08{\tiny$\pm$6.29} &96.71{\tiny$\pm$0.90} &\underline{92.69{\tiny$\pm$1.59}} & 92.19{\tiny$\pm$2.06} &53.43{\tiny$\pm$2.10} &44.86{\tiny$\pm$1.11} &35.70{\tiny$\pm$1.22} &98.48{\tiny$\pm$0.17} &96.37{\tiny$\pm$0.23} &96.66{\tiny$\pm$0.35} \\

(AAAI'24) &ICMVC 
&83.97{\tiny$\pm$7.69} &80.72{\tiny$\pm$5.65} &74.32{\tiny$\pm$9.42} &89.17{\tiny$\pm$7.27} &83.83{\tiny$\pm$8.27} &82.22{\tiny$\pm$7.39} &41.20{\tiny$\pm$2.19} &32.00{\tiny$\pm$1.01} &24.56{\tiny$\pm$1.95} &99.32{\tiny$\pm$0.05} &98.04{\tiny$\pm$0.12} &98.50{\tiny$\pm$0.11} \\

(AAAI'24) &SURER 
& 86.54{\tiny$\pm$4.60} &82.36{\tiny$\pm$2.33} &75.88{\tiny$\pm$5.64} &84.06{\tiny$\pm$3.71} &74.12{\tiny$\pm$2.22} &69.37{\tiny$\pm$3.92} &54.95{\tiny$\pm$0.87} &51.37{\tiny$\pm$0.92} &40.55{\tiny$\pm$0.99} &83.16{\tiny$\pm$1.71} &79.72{\tiny$\pm$1.76} &74.51{\tiny$\pm$1.87} \\

(CVPR'24) &MVCAN 
&75.86{\tiny$\pm$1.44} &77.51{\tiny$\pm$1.61} &68.81{\tiny$\pm$2.27} &\underline{96.81{\tiny$\pm$1.09}} &92.02{\tiny$\pm$2.34} &\underline{92.36{\tiny$\pm$2.59}} &55.04{\tiny$\pm$1.04} &\textbf{54.60{\tiny$\pm$0.79}} &\underline{42.79{\tiny$\pm$1.13}} &82.42{\tiny$\pm$4.29} &81.11{\tiny$\pm$1.67} &75.26{\tiny$\pm$3.20} \\

\midrule

(NIPS'20) &PVC 
&90.12{\tiny$\pm$0.17} &80.90{\tiny$\pm$0.28} &79.58{\tiny$\pm$0.34} &95.86{\tiny$\pm$0.02} &90.61{\tiny$\pm$0.05} &90.17{\tiny$\pm$0.05} &44.31{\tiny$\pm$0.14} &46.14{\tiny$\pm$0.46} &30.16{\tiny$\pm$0.19} &89.11{\tiny$\pm$0.07} &80.10{\tiny$\pm$0.08} &77.55{\tiny$\pm$0.13} \\

(CVPR'21) & MvCLN 
&82.31{\tiny$\pm$1.26} &73.96{\tiny$\pm$0.91} &66.81{\tiny$\pm$1.64} &81.83{\tiny$\pm$2.46} &70.64{\tiny$\pm$1.27} &65.26{\tiny$\pm$2.34} &53.20{\tiny$\pm$0.31} &44.60{\tiny$\pm$1.44} &32.52{\tiny$\pm$0.74} &99.24{\tiny$\pm$0.02} &97.80{\tiny$\pm$0.06} &98.32{\tiny$\pm$0.05} \\

(TPAMI'23) &SURE 
&72.62{\tiny$\pm$1.09} &68.94{\tiny$\pm$1.07} &57.51{\tiny$\pm$1.48} & 78.24{\tiny$\pm$5.38} &63.47{\tiny$\pm$3.59} & 58.41{\tiny$\pm$7.52} &56.27{\tiny$\pm$0.14} & 46.96{\tiny$\pm$1.15} &35.91{\tiny$\pm$0.29} & 99.15{\tiny$\pm$0.02} &97.61{\tiny$\pm$0.05} & 98.13{\tiny$\pm$0.04} \\

(MM'24) & CGDA 
&90.06{\tiny$\pm$1.23} & 83.52{\tiny$\pm$1.65} &80.29{\tiny$\pm$1.85} & 91.44{\tiny$\pm$2.70} &82.25{\tiny$\pm$2.46} & 80.82{\tiny$\pm$3.72} &\underline{56.80{\tiny$\pm$0.12}} & 48.91{\tiny$\pm$0.13} &40.29{\tiny$\pm$0.04} & 97.48{\tiny$\pm$0.88} &94.19{\tiny$\pm$0.67} & 94.50{\tiny$\pm$0.63} \\

(MM'24) &VITAL 
&\underline{92.23{\tiny$\pm$0.19}} &\underline{85.12{\tiny$\pm$0.35}} &\underline{83.39{\tiny$\pm$0.32}} &85.37{\tiny$\pm$4.83} &76.87{\tiny$\pm$5.12} & 70.03{\tiny$\pm$7.15} &55.03{\tiny$\pm$0.51} & 53.58{\tiny$\pm$0.54} &41.81{\tiny$\pm$0.81} &\textbf{99.80{\tiny$\pm$0.06}} &\textbf{99.40{\tiny$\pm$0.14}} &\textbf{99.55{\tiny$\pm$0.12}} \\

(TCSVT'24) &CGCN 
&73.57{\tiny$\pm$1.15} &73.19{\tiny$\pm$1.95} &64.26{\tiny$\pm$2.69} &96.77{\tiny$\pm$1.57} &91.20{\tiny$\pm$2.26} &92.40{\tiny$\pm$3.38} &54.57{\tiny$\pm$0.57} &51.81{\tiny$\pm$0.57} &41.58{\tiny$\pm$0.74} &98.55{\tiny$\pm$0.03} &96.16{\tiny$\pm$0.04} &96.81{\tiny$\pm$0.07} \\ 

(TCSVT'25) &TCLPVC  
&\underline{93.09{\tiny$\pm$1.37}} &\underline{86.71{\tiny$\pm$1.58}} &\underline{85.09{\tiny$\pm$2.37}} &\underline{97.13{\tiny$\pm$0.94}} &\underline{93.92{\tiny$\pm$1.31}} &\underline{94.17{\tiny$\pm$1.57}} &48.64{\tiny$\pm$2.32} &45.06{\tiny$\pm$1.98} &33.15{\tiny$\pm$2.74} &98.50{\tiny$\pm$0.13} &95.98{\tiny$\pm$0.32} &96.69{\tiny$\pm$0.27} \\ 

\rowcolor{green!10} (\textbf{Ours}) & \textbf{SMART} & \textbf{93.21{\tiny$\pm$0.39}} & \textbf{86.78{\tiny$\pm$0.62}} & \textbf{85.56{\tiny$\pm$0.77}} & \textbf{97.95{\tiny$\pm$0.10}} & \textbf{94.60{\tiny$\pm$0.35}} & \textbf{95.00{\tiny$\pm$0.24}} & \textbf{60.33{\tiny$\pm$1.73}} & \underline{53.35{\tiny$\pm$0.93}} & \textbf{44.09{\tiny$\pm$1.40}} & \underline{99.55{\tiny$\pm$0.19}} & \underline{98.73{\tiny$\pm$0.46}} & \underline{99.00{\tiny$\pm$0.41}} \\
\midrule[0.85pt]
\multicolumn{2}{c|}{\multirow{2}{*}{\textbf{Methods}}} & \multicolumn{3}{c|}{\textbf{NUS-WIDE}} & \multicolumn{3}{c|}{\textbf{Hdigit}} & \multicolumn{3}{c|}{\textbf{Deep Animal}} & \multicolumn{3}{c}{\textbf{Reuters}} \\

& &\textbf{ACC} & \textbf{NMI} & \textbf{ARI} & \textbf{ACC} & \textbf{NMI} & \textbf{ARI} & \textbf{ACC} & \textbf{NMI} & \textbf{ARI} & \textbf{ACC} & \textbf{NMI} & \textbf{ARI} \\
\midrule
(CVPR'22)& MFLVC 
&27.12{\tiny$\pm$0.66} &18.28{\tiny$\pm$0.85} &12.34{\tiny$\pm$0.67} &\textbf{99.79{\tiny$\pm$0.01}} &97.61{\tiny$\pm$0.11} &\textbf{99.52{\tiny$\pm$0.02}} &33.50{\tiny$\pm$1.54} &44.80{\tiny$\pm$0.99} &21.40{\tiny$\pm$0.88} &47.75{\tiny$\pm$2.59} &29.38{\tiny$\pm$1.87} &20.24{\tiny$\pm$1.59} \\

(MM'23) & DealMVC 
&47.36{\tiny$\pm$9.68} &33.71{\tiny$\pm$7.34} &27.60{\tiny$\pm$7.48} &91.16{\tiny$\pm$1.48} &80.74{\tiny$\pm$2.14} &81.73{\tiny$\pm$2.69} &37.16{\tiny$\pm$0.77} &50.47{\tiny$\pm$0.71} &27.77{\tiny$\pm$0.62} &52.07{\tiny$\pm$4.51} &33.55{\tiny$\pm$3.62} &27.86{\tiny$\pm$4.01} \\

(CVPR'23) & GCFAgg 
&36.22{\tiny$\pm$1.01} &27.03{\tiny$\pm$2.06} &18.64{\tiny$\pm$1.26} &96.43{\tiny$\pm$3.71} &94.86{\tiny$\pm$2.40} &93.99{\tiny$\pm$4.52} &19.92{\tiny$\pm$0.65} &32.49{\tiny$\pm$1.54} &13.66{\tiny$\pm$1.32} &48.46{\tiny$\pm$3.62} &32.71{\tiny$\pm$1.58} &26.72{\tiny$\pm$3.27}  \\

(AAAI'24) & ICMVC 
&65.76{\tiny$\pm$1.38} &51.63{\tiny$\pm$0.85} &47.03{\tiny$\pm$1.55} &99.35{\tiny$\pm$0.11} &97.96{\tiny$\pm$0.29} &98.55{\tiny$\pm$0.23} &47.16{\tiny$\pm$0.65} &62.85{\tiny$\pm$0.23} &36.90{\tiny$\pm$0.68} &49.13{\tiny$\pm$2.57} &33.61{\tiny$\pm$2.73} &25.36{\tiny$\pm$2.81} \\

(AAAI'24) & SURER 
&32.40{\tiny$\pm$6.31} &25.79{\tiny$\pm$7.96} &11.81{\tiny$\pm$4.99} &85.92{\tiny$\pm$6.99} &79.42{\tiny$\pm$3.75} &74.13{\tiny$\pm$10.10} &21.06{\tiny$\pm$1.05} &32.47{\tiny$\pm$1.13} &10.37{\tiny$\pm$1.18} &47.61{\tiny$\pm$2.59} &16.82{\tiny$\pm$3.61} &15.09{\tiny$\pm$3.79} \\

(CVPR'24) & MVCAN 
&41.91{\tiny$\pm$4.52} &34.48{\tiny$\pm$1.97} &24.91{\tiny$\pm$2.55} &95.80{\tiny$\pm$0.40} &90.27{\tiny$\pm$0.58} &90.98{\tiny$\pm$0.80} &\underline{58.93{\tiny$\pm$1.03}} &\textbf{70.71{\tiny$\pm$0.89}} &\underline{51.70{\tiny$\pm$1.04}} &48.53{\tiny$\pm$3.05} &26.03{\tiny$\pm$4.27} &25.58{\tiny$\pm$4.47} \\

\midrule

(NIPS'20) & PVC 
&52.12{\tiny$\pm$0.96} &42.16{\tiny$\pm$0.73} &37.94{\tiny$\pm$0.68} &87.45{\tiny$\pm$0.01} &92.49{\tiny$\pm$0.03} &93.53{\tiny$\pm$0.03} &3.83{\tiny$\pm$0.00} &0.00{\tiny$\pm$0.00} &0.00{\tiny$\pm$0.00} &40.73{\tiny$\pm$0.51} &20.52{\tiny$\pm$0.41} &15.11{\tiny$\pm$0.38} \\

(CVPR'21) & MvCLN 
&62.58{\tiny$\pm$0.72} &46.64{\tiny$\pm$1.15} &41.59{\tiny$\pm$0.98} &98.61{\tiny$\pm$0.11} &96.09{\tiny$\pm$0.24} &96.95{\tiny$\pm$0.24} &40.11{\tiny$\pm$0.43} &53.59{\tiny$\pm$0.68} &32.28{\tiny$\pm$0.79} &50.82{\tiny$\pm$1.59} &30.01{\tiny$\pm$1.63} &25.72{\tiny$\pm$1.83} \\

(TPAMI'23) & SURE 
&63.49{\tiny$\pm$0.23} &49.58{\tiny$\pm$0.16} &44.87{\tiny$\pm$0.27} &97.96{\tiny$\pm$0.02} &94.31{\tiny$\pm$0.07} &95.51{\tiny$\pm$0.05} &41.44{\tiny$\pm$0.08} &54.92{\tiny$\pm$0.21} &32.13{\tiny$\pm$0.23} &50.37{\tiny$\pm$1.52} &29.25{\tiny$\pm$2.17} &24.39{\tiny$\pm$1.92} \\

(MM'24) & CGDA 
&47.92{\tiny$\pm$2.74} &43.23{\tiny$\pm$3.10} &26.32{\tiny$\pm$2.68} &96.91{\tiny$\pm$0.54} &92.80{\tiny$\pm$1.07} &93.29{\tiny$\pm$1.14} &25.44{\tiny$\pm$1.06} &41.90{\tiny$\pm$0.90} &18.34{\tiny$\pm$0.78} &\underline{56.25{\tiny$\pm$0.28}} &34.09{\tiny$\pm$0.12} &28.64{\tiny$\pm$0.29} \\

(MM'24) & VITAL 
&\underline{66.07{\tiny$\pm$1.28}} &\textbf{54.02{\tiny$\pm$0.59}} &\underline{48.22{\tiny$\pm$1.31}} &\underline{99.75{\tiny$\pm$0.03}} &\textbf{99.19{\tiny$\pm$0.07}} &\underline{99.44{\tiny$\pm$0.05}} &55.56{\tiny$\pm$1.41} &64.64{\tiny$\pm$0.47} &44.48{\tiny$\pm$1.40} &50.04{\tiny$\pm$3.45} &34.82{\tiny$\pm$2.64} &25.52{\tiny$\pm$5.64} \\

(TCSVT'24) &CGCN 
&60.79{\tiny$\pm$0.42} &47.60{\tiny$\pm$0.40} &40.50{\tiny$\pm$0.41} &98.95{\tiny$\pm$0.03} &96.94{\tiny$\pm$0.09} &97.69{\tiny$\pm$0.07} &43.99{\tiny$\pm$2.36} &55.02{\tiny$\pm$2.42} &31.79{\tiny$\pm$2.27} &46.53{\tiny$\pm$0.84} &28.61{\tiny$\pm$1.20} &23.41{\tiny$\pm$0.89} \\

(TCSVT'25) &TCLPVC 
&46.40{\tiny$\pm$1.72} &33.61{\tiny$\pm$1.11} &26.36{\tiny$\pm$1.20} &99.26{\tiny$\pm$0.08} &97.85{\tiny$\pm$0.20} &98.37{\tiny$\pm$0.18} &50.64{\tiny$\pm$1.51} &67.18{\tiny$\pm$0.96} &43.02{\tiny$\pm$2.20} &54.92{\tiny$\pm$2.55} &\underline{36.47{\tiny$\pm$2.84}} &\underline{30.07{\tiny$\pm$2.15}} \\

\rowcolor{green!10} (\textbf{Ours}) &\textbf{SMART} 
&\textbf{67.25{\tiny$\pm$0.24}} &\underline{53.58{\tiny$\pm$0.21}} &\textbf{48.35{\tiny$\pm$0.33}} &99.66{\tiny$\pm$0.07} &\underline{98.96{\tiny$\pm$0.17}} &99.24{\tiny$\pm$0.14} &\textbf{63.47{\tiny$\pm$0.99}} &\underline{69.28{\tiny$\pm$0.68}} &\textbf{52.33{\tiny$\pm$0.50}} &\textbf{59.14{\tiny$\pm$1.08}} &\textbf{36.74{\tiny$\pm$0.97}} &\textbf{35.35{\tiny$\pm$1.40}} \\ 
\bottomrule[1.25pt]
\end{tabular}
}
\label{tab:methods_comparison_100}
\end{table*}
\subsection{Optimization}
The optimization of SMART involves minimizing a composite loss function that balances reconstruction fidelity, view distribution alignment, and semantic matching contrastive learning. The total loss is defined as:
\begin{equation}\label{Eq:loss_total}
    \mathcal{L} = \mathcal{L}_{rec} + \lambda_1 \mathcal{L}_{vda} + \lambda_2 \mathcal{L}_{smc},
\end{equation}
where $\lambda_1$ and $\lambda_2$ are two weighting factors which control the trade-off between these loss terms, and their values are analyzed in Section \ref{sec:parameter}.
As described in Algorithm \ref{algorithm}, our proposed SMART adopts a single-stage training paradigm without requiring time-consuming pretraining. In contrast to existing PVC methods that rely on correspondence learning to perform instance-level or category-level realignment, our approach directly leverages the common semantic representations optimized via semantic matching contrastive learning for downstream tasks. This design avoids the potential negative effects caused by erroneous realignment, thereby improving both efficiency and robustness.

\subsection{Computational Complexity Analysis.}
The computational complexity of our proposed method can be analyzed from two perspectives: the entire networks (comprising the autoencoders and the high-level projector) and the three proposed objectives. We adopt the mini-batch strategy for model training, and set the batch size to $B$, where $B \le N$. The autoencoders exhibit complexity of $\mathcal{O}(VND_v)$, while the high-level projector requires $\mathcal{O}(VNd^2)$  complexity. Here we have $D_v > d$. 
Regarding the objective functions, the computational cost primarily stems from the calculation of $\mathcal{L}_{rec}$, $\mathcal{L}_{vda}$, and $\mathcal{L}_{smc}$. Their complexities are $\mathcal{O}(VND_v)$, $\mathcal{O}(V(N/B)B_a^2d)$, and $\mathcal{O}(VNBd)$ respectively, where $B_a \le B$ denotes the number of aligned samples in a mini-batch and thus we have $\mathcal{O}(V(N/B)B_a^2d) \le \mathcal{O}(VNBd)$. 
Generally, it holds that $B > D_v$. 
Consequently, the highest complexity order of our method is $\mathcal{O}(VNBd)$, which aligns with that of the classical InfoNCE \cite{2018CPC-oord} objective. This design ensures that our method maintains performance advantages without exceeding conventional contrastive learning frameworks' cost.

\begin{figure*}[!t]
    \centering
    \includegraphics[width=1.0\linewidth]{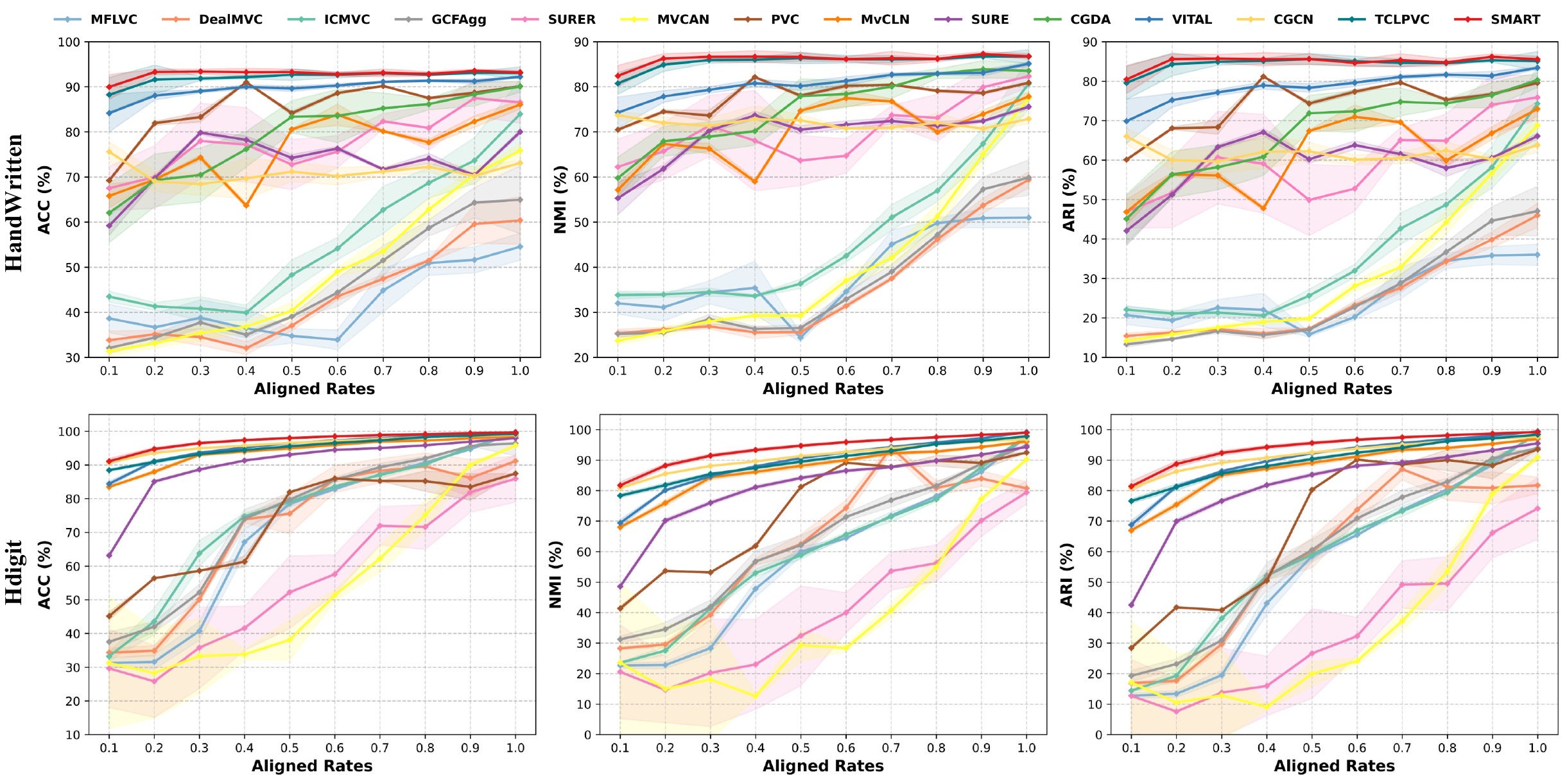}
    \caption{Clustering performance on HandWritten and Hdigit datasets across varying alignment rates (mean $\pm$ std. over 5 runs). Solid lines show the mean values while shaded regions indicate standard deviation. (CGDA encountered out-of-memory error ``24GB GPU limit" on Hdigit below 30\% alignment).}
    \label{fig:Align_diffalign}
\end{figure*}
\section{Experiments}
\subsection{Experimental Setups}
\label{sub:Experimental Setups}

\noindent \textbf{Datasets and Evaluation Metrics.} 
We conduct comprehensive evaluations on eight benchmark multi-view datasets: HandWritten, BDGP, WIKI, MNIST-USPS, NUS-WIDE, Reuters, Hdigit, and Deep Animal. To assess robustness, we train the model under both fully aligned and partially aligned scenarios. For partial alignment, we simulate real-world conditions by randomly shuffling instance pairs, with alignment rate $\eta = \frac{N_{a}}{N}$. Table \ref{tab:dataset_info} summarizes key dataset statistics. 
Clustering performance is quantified using three widely used metrics: clustering Accuracy (ACC), Normalized Mutual Information (NMI), and Adjusted Rand Index (ARI). Higher values indicate better performance across all metrics.

\noindent \textbf{Competing methods.} We compare SMART against several state-of-the-art methods, including general multi-view clustering: MFLVC \cite{2022MFLVC-xu}, DealMVC \cite{2023DealMVC-yang}, GCFAgg \cite{2023GCFAgg-yan}, ICMVC \cite{2024ICMVC-chao}, Surer \cite{2024SURE-wang}, MVCAN \cite{2024MVCAN-Xu}, and PVC-specific methods: PVC \cite{2020PVC-huang}, MvCLN\cite{2021MvCLN-yang}, SURE \cite{2022SURE-yang}, CGDA \cite{2024CGDA-wang}, VITAL \cite{2024VITAL-he}, CGCN \cite{2024CGCN-wang}, and TCLPVC \cite{2025TCLPVC-gao}. For fair comparison with general MVC methods in partial alignment scenarios, we preprocess data using PCA for dimensionality reduction followed by instance re-pairing via the Hungarian algorithm.

\noindent \textbf{Implementation Details.}
All experiments are conducted on Ubuntu 18.04 (NVIDIA 4090 GPU) using PyTorch 1.12.1. The model is trained for 500 epochs until convergence using the Adam \cite{2017Adam-Kingma} optimizer with an initial learning rate of 1e-4. The training strategy involves joint optimization of all objectives. The temperature parameter $\tau$ is set to 1.0 by default. 

\subsection{Comparative Analysis}
\label{sub:Comparative Analysis}

\begin{table*}[h]
    \renewcommand{\arraystretch}{1.1}
    \centering
    \caption{Ablation studies of our method. The best results are indicated in bold.}
    \scalebox{0.82}{
    \fontsize{9.5}{11}\selectfont 
    \setlength{\tabcolsep}{2.5pt}
    \begin{tabular}{c|l|ccc|ccc|ccc|ccc}
	\toprule[1.25pt]
        \multirow{2}{*}{\textbf{Aligned}} & \multicolumn{1}{c|}{\multirow{2}{*}{\textbf{Variants}}} & \multicolumn{3}{c|}{\textbf{HandWritten}} & \multicolumn{3}{c|}{\textbf{BDGP}} & \multicolumn{3}{c|}{\textbf{MNIST-USPS}}& \multicolumn{3}{c}{\textbf{Hdigit}}\\
        & & \textbf{ACC} & \textbf{NMI} & \textbf{ARI} & \textbf{ACC} & \textbf{NMI} & \textbf{ARI} & \textbf{ACC} & \textbf{NMI} & \textbf{ARI} & \textbf{ACC} & \textbf{NMI} & \textbf{ARI} \\
	\midrule
        \multirow{4}{*}{\text{Partially}} \  
        & ~ $\mathcal{L}_{rec}$ \ &71.56{\tiny$\pm$3.80} & 60.72{\tiny$\pm$3.78} & 50.61{\tiny$\pm$5.15} & 82.16{\tiny$\pm$3.05} & 64.21{\tiny$\pm$2.62} & 61.14{\tiny$\pm$5.55} & 52.00{\tiny$\pm$3.39} & 44.38{\tiny$\pm$2.81} & 32.70{\tiny$\pm$2.68} & 51.05{\tiny$\pm$3.15} & 39.81{\tiny$\pm$3.64} & 27.97{\tiny$\pm$2.98}\\
        & ~ $\mathcal{L}_{rec} + \mathcal{L}_{vda}$ \ & 73.32{\tiny$\pm$0.73} & 67.01{\tiny$\pm$1.69} & 55.90{\tiny$\pm$1.76} & 88.36{\tiny$\pm$4.17} & 77.41{\tiny$\pm$6.90} & 74.44{\tiny$\pm$8.58} & 90.18{\tiny$\pm$1.67} & 81.80{\tiny$\pm$1.73} & 80.20{\tiny$\pm$2.78} & 95.75{\tiny$\pm$0.65} & 90.29{\tiny$\pm$1.03} & 90.87{\tiny$\pm$1.34}\\
        & ~ $\mathcal{L}_{rec} + \mathcal{L}_{vda} + \mathcal{L}_{smc}$ ~ 
        &\textbf{93.27}{\tiny$\pm$0.67} & \textbf{86.62}{\tiny$\pm$0.96} 
        & \textbf{85.62}{\tiny$\pm$1.34} & \textbf{95.58}{\tiny$\pm$0.52} 
        & \textbf{89.44}{\tiny$\pm$1.48} & \textbf{89.47}{\tiny$\pm$1.22} 
        & \textbf{95.77}{\tiny$\pm$0.72} & \textbf{89.99}{\tiny$\pm$1.02} 
        & \textbf{90.91}{\tiny$\pm$1.44} & \textbf{98.00}{\tiny$\pm$0.24} 
        & \textbf{94.73}{\tiny$\pm$0.44} & \textbf{95.61}{\tiny$\pm$0.51}\\
        & ~ $w/o$ Guidance 
        & 91.74{\tiny$\pm$1.28} & 84.57{\tiny$\pm$1.94} & 82.83{\tiny$\pm$2.50} & 92.10{\tiny$\pm$1.02} & 81.81{\tiny$\pm$2.69} & 82.17{\tiny$\pm$2.66} & 93.76{\tiny$\pm$0.63} & 86.48{\tiny$\pm$1.11} & 86.75{\tiny$\pm$1.28} & 96.37{\tiny$\pm$0.25} & 91.10{\tiny$\pm$0.46} & 92.13{\tiny$\pm$0.52}\\
        \midrule
        \multirow{4}{*}{Fully} & ~ $\mathcal{L}_{rec}$ & 71.00{\tiny$\pm$4.00} & 60.46{\tiny$\pm$3.08} & 49.83{\tiny$\pm$5.57} & 82.11{\tiny$\pm$3.46} & 65.71{\tiny$\pm$3.97} & 61.40{\tiny$\pm$6.56} & 52.06{\tiny$\pm$3.32} & 44.20{\tiny$\pm$3.31} & 32.66{\tiny$\pm$3.00} & 50.76{\tiny$\pm$3.17} & 39.63{\tiny$\pm$4.10} & 27.94{\tiny$\pm$3.28}\\
	& ~ $\mathcal{L}_{rec} + \mathcal{L}_{vda}$
        & 72.85{\tiny$\pm$0.63} & 65.83{\tiny$\pm$2.51} & 55.45{\tiny$\pm$1.33} & 96.65{\tiny$\pm$0.88} & 91.55{\tiny$\pm$1.57} & 91.93{\tiny$\pm$2.02} & 93.69{\tiny$\pm$0.81} & 88.10{\tiny$\pm$1.12} & 86.91{\tiny$\pm$1.55} & 97.31{\tiny$\pm$0.44} & 93.96{\tiny$\pm$0.70} & 94.19{\tiny$\pm$0.92}\\
	& ~ $\mathcal{L}_{rec} + \mathcal{L}_{vda} + \mathcal{L}_{smc}$ ~
        & \textbf{93.21}{\tiny$\pm$0.39} & \textbf{86.78}{\tiny$\pm$0.62} 
        & \textbf{85.56}{\tiny$\pm$0.77} & \textbf{97.95}{\tiny$\pm$0.10} 
        & \textbf{94.60}{\tiny$\pm$0.35} & \textbf{95.00}{\tiny$\pm$0.24} 
        & \textbf{99.55}{\tiny$\pm$0.19} & \textbf{98.73}{\tiny$\pm$0.46} 
        & \textbf{99.00}{\tiny$\pm$0.41} & \textbf{99.66}{\tiny$\pm$0.07} 
        & \textbf{98.96}{\tiny$\pm$0.17} & \textbf{99.24}{\tiny$\pm$0.14}\\
        & ~ $w/o$ Guidance 
        & 91.99{\tiny$\pm$0.88} & 84.79{\tiny$\pm$1.06} & 83.21{\tiny$\pm$1.66} & 96.38{\tiny$\pm$1.68} & 90.55{\tiny$\pm$3.26} & 91.26{\tiny$\pm$3.89} & 97.43{\tiny$\pm$0.60} & 94.58{\tiny$\pm$0.99} & 94.48{\tiny$\pm$1.24} & 98.57{\tiny$\pm$0.24} & 97.12{\tiny$\pm$0.58} & 97.73{\tiny$\pm$0.51}\\
	\bottomrule[1.25pt]
    \end{tabular}
    }
    \label{tab:ablation}
\end{table*}
\begin{table*}[t]
    \renewcommand\arraystretch{1.0}
    \centering
    \caption{Ablation studies of our method. The best results are indicated in bold.}
    \scalebox{0.86}{
    \fontsize{9.5}{11}\selectfont 
    \setlength{\tabcolsep}{1.0mm}
    \begin{tabular}{c|l|ccc|ccc|ccc|ccc}
	\toprule[1.25pt]
        \multirow{2}{*}{\textbf{Aligned}} & \multicolumn{1}{c|}{\multirow{2}{*}{\textbf{Variants}}} & \multicolumn{3}{c|}{\textbf{HandWritten}} & \multicolumn{3}{c|}{\textbf{BDGP}} & \multicolumn{3}{c|}{\textbf{MNIST-USPS}}& \multicolumn{3}{c}{\textbf{Hdigit}}\\
        & & \textbf{ACC} & \textbf{NMI} & \textbf{ARI} 
        & \textbf{ACC} & \textbf{NMI} & \textbf{ARI} 
        & \textbf{ACC} & \textbf{NMI} & \textbf{ARI} 
        & \textbf{ACC} & \textbf{NMI} & \textbf{ARI} \\
	\midrule
        \multirow{4}{*}{\text{Partially}} ~~ 
        & ~ $w/o$ $\mathcal{L}_{cfa}$ \ & 75.79{\tiny$\pm$4.03} & 67.57{\tiny$\pm$2.74} & 58.23{\tiny$\pm$4.25} & 65.49{\tiny$\pm$6.83} & 42.23{\tiny$\pm$7.16} & 35.08{\tiny$\pm$8.86} & 40.47{\tiny$\pm$1.42} & 30.14{\tiny$\pm$2.61} & 20.46{\tiny$\pm$1.92} & 31.87{\tiny$\pm$1.85} & 20.94{\tiny$\pm$3.34} & 12.46{\tiny$\pm$1.16} \\
        & ~ $w/o$ $\mathcal{L}_{cma}$ \ & 89.45{\tiny$\pm$3.39} & 85.76{\tiny$\pm$1.89} & 82.24{\tiny$\pm$3.37} & 87.06{\tiny$\pm$6.49} & 79.51{\tiny$\pm$6.05} & 74.85{\tiny$\pm$10.10} & 74.02{\tiny$\pm$6.46} & 72.35{\tiny$\pm$5.14} & 62.95{\tiny$\pm$7.37} & 89.08{\tiny$\pm$3.00} & 80.93{\tiny$\pm$3.81} & 78.04{\tiny$\pm$5.31} \\
        & ~ Ours ~ 
        &\textbf{93.27}{\tiny$\pm$0.67} & \textbf{86.62}{\tiny$\pm$0.96} 
        & \textbf{85.62}{\tiny$\pm$1.34} & \textbf{95.58}{\tiny$\pm$0.52} 
        & \textbf{89.44}{\tiny$\pm$1.48} & \textbf{89.47}{\tiny$\pm$1.22} 
        & \textbf{95.77}{\tiny$\pm$0.72} & \textbf{89.99}{\tiny$\pm$1.02} 
        & \textbf{90.91}{\tiny$\pm$1.44} & \textbf{98.00}{\tiny$\pm$0.24} 
        & \textbf{94.73}{\tiny$\pm$0.44} & \textbf{95.61}{\tiny$\pm$0.51}\\
        \midrule
        \multirow{4}{*}{Fully} 
        & ~ $w/o$ $\mathcal{L}_{cfa}$
        & 74.93{\tiny$\pm$2.06} & 66.39{\tiny$\pm$2.04} & 57.55{\tiny$\pm$2.47} & 61.30{\tiny$\pm$5.81} & 43.28{\tiny$\pm$8.48} & 33.02{\tiny$\pm$5.70} & 40.65{\tiny$\pm$3.04} & 29.32{\tiny$\pm$2.75} & 18.80{\tiny$\pm$2.72} & 33.97{\tiny$\pm$0.78} & 24.77{\tiny$\pm$1.72} & 14.32{\tiny$\pm$0.59} \\
        & ~ $w/o$ $\mathcal{L}_{cma}$ & 87.92{\tiny$\pm$1.79} & 85.13{\tiny$\pm$1.26} & 80.30{\tiny$\pm$1.60} & 90.74{\tiny$\pm$7.36} & 85.31{\tiny$\pm$7.58} & 82.70{\tiny$\pm$10.27} & 72.51{\tiny$\pm$5.61} & 68.99{\tiny$\pm$2.75} & 58.10{\tiny$\pm$4.21} & 89.91{\tiny$\pm$2.77} & 83.42{\tiny$\pm$3.71} & 79.92{\tiny$\pm$4.88} \\
	& ~ Ours ~
        & \textbf{93.21}{\tiny$\pm$0.39} & \textbf{86.78}{\tiny$\pm$0.62} 
        & \textbf{85.56}{\tiny$\pm$0.77} & \textbf{97.95}{\tiny$\pm$0.10} 
        & \textbf{94.60}{\tiny$\pm$0.35} & \textbf{95.00}{\tiny$\pm$0.24} 
        & \textbf{99.55}{\tiny$\pm$0.19} & \textbf{98.73}{\tiny$\pm$0.46} 
        & \textbf{99.00}{\tiny$\pm$0.41} & \textbf{99.66}{\tiny$\pm$0.07} 
        & \textbf{98.96}{\tiny$\pm$0.17} & \textbf{99.24}{\tiny$\pm$0.14}\\
	\bottomrule[1.25pt]
    \end{tabular}
    }
    \label{tab:ablation_cov}
\end{table*}

To comprehensively evaluate the effectiveness of SMART, we conduct a series of comparative experiments under varying view alignment conditions: partial alignment (50\%) in Table \ref{tab:methods_comparison_50}, full alignment (100\%) in Table \ref{tab:methods_comparison_100}. Each method is executed five times with mean values reported alongside corresponding standard deviations across eight benchmark datasets to ensure statistical reliability.
From these experimental results, we derive the following observations:
\begin{itemize}
    \item In the partially aligned (50\%) setting, SMART consistently outperforms both PVC and traditional MVC approaches, across all datasets and metrics. Notably, even on multi-modal datasets with significant inter-view distribution disparities, such as BDGP, WIKI, and NUS-WIDE, SMART achieves substantial performance gains. This superiority is primarily attributed to its effective alignment of view distributions and the use of semantic graph to enable comprehensive representation learning across all samples, resulting in more robust clustering outcomes. 
    \item Under the fully aligned (100\%) scenario, SMART maintains its lead across all datasets even against generic MVC methods, delivering the best results on most benchmarks, e.g., HandWritten, BDGP, Reuters, and others. On MNIST-USPS (ACC=99.55\%) and Hdigit (ACC=99.66\%), it achieves comparable performance, with no statistically significant difference from the best method. The superiority owes to its ability to capture common semantic information, which accordingly leads to more discriminative cluster assignments. 
    \item SMART’s exceptional performance in both settings stems from its ability to learn consistent features from both aligned and unaligned data, mitigating the degradation of representation quality in partial alignment scenarios. This robustness underscores its broad applicability to diverse multi-view clustering challenges. 
\end{itemize}

\subsection{Model Analysis}
\subsubsection{\textbf{Ablation Studies}}
\label{sub:Ablation Studies}
To assess the contributions of the view distribution alignment ($\mathcal{L}_{vda}$) and semantic matching contrastive learning ($\mathcal{L}_{smc}$), we conduct systematic ablation studies recorded in Table \ref{tab:ablation}. From the results, we can observe that solely adopting intra-view feature reconstruction yields poor clustering. While by incorporating $\mathcal{L}_{vda}$ and $\mathcal{L}_{smc}$, our approach achieves varying degrees of improvement in clustering metrics. Notably, in partial alignment, removing $\mathcal{L}_{smc}$ (training only on aligned data) markedly reduces performance. This enhancement stems from the reduction of cross-view feature distribution discrepancies and the capture of common semantics from both aligned and unaligned data, which collectively elevate the quality of representations and lead to superior clustering results. 

In addition, to dissect the contributions of two sub-components cross-view feature alignment loss $\mathcal{L}_{cfa}$ and covariance matching alignment loss $\mathcal{L}_{cma}$ comprised within the view distribution alignment loss $\mathcal{L}_{vda}$, we perform ablation studies by individually removing each term and evaluating the impact on clustering performance across four datasets.
The results recorded in Table \ref{tab:ablation_cov} show that removing $\mathcal{L}_{cfa}$ leads to a significant drop in performance, indicating its critical role in aligning cross-view distributions. Further, omitting $\mathcal{L}_{cma}$ also degrades results, though to a lesser extent, suggesting it aids in stabilizing the latent space. The full model consistently outperforms ablated variants, confirming the synergy of both sub-components in $\mathcal{L}_{vda}$.

\subsubsection{\textbf{Does Semantic Guidance Graph Truly Matter}}
To evaluate the role of semantic guidance graph, we compare SMART against a variant without guidance ($w/o$ Guidance), as shown in Table \ref{tab:ablation}. The variant omits semantic relationships, relying only on positive and negative pairs and training solely on aligned data. Moreover, we visualize the learned semantic graphs as heatmaps in Fig. \ref{fig:adj}, revealing strong alignment with ground-truth category matrices, with high correlations in true class regions. Evidently, removing the semantic graph results in performance degradation in both partially (50\%) and fully (100\%) aligned settings, confirming its critical role in our approach. These results demonstrate that the semantic graph guides contrastive learning to produce discriminative, consistent representations across aligned and unaligned views, enhancing clustering reliability.

\begin{table*}[t]
    \renewcommand\arraystretch{1.0}
    \centering
    \caption{Clustering performance (\%) of PVC-specific methods against completely unaligned MVC approaches (MVC-UM and OpVuC) on extreme alignment (1\%) scenario. Best results are marked in bold. `-' indicates unavailable result, while `OOM' denotes out-of-memory errors on a 24 GB GPU.}
    \scalebox{0.86}{
    \fontsize{9.5}{11}\selectfont 
    \setlength{\tabcolsep}{4.8pt}
    \begin{tabular}{c|ccc|ccc|ccc|ccc}
	\toprule[1.25pt]
        \multicolumn{1}{c|}{\multirow{2}{*}{\textbf{Methods}}} & \multicolumn{3}{c|}{\textbf{HandWritten}} & \multicolumn{3}{c|}{\textbf{BDGP}} & \multicolumn{3}{c|}{\textbf{MNIST-USPS}} & \multicolumn{3}{c}{\textbf{Hdigit}} \\
        & \textbf{ACC} & \textbf{NMI} & \textbf{ARI} 
        & \textbf{ACC} & \textbf{NMI} & \textbf{ARI} 
        & \textbf{ACC} & \textbf{NMI} & \textbf{ARI} 
        & \textbf{ACC} & \textbf{NMI} & \textbf{ARI} \\
	\midrule
        MVC-UM
        &66.53{\tiny$\pm$2.15} &61.27{\tiny$\pm$3.44} &\textbf{52.01{\tiny$\pm$2.36}} &42.08{\tiny$\pm$3.47} &20.88{\tiny$\pm$4.24} &15.34{\tiny$\pm$4.57} &47.59{\tiny$\pm$2.07} &\textbf{40.12{\tiny$\pm$2.73}} &\textbf{28.73{\tiny$\pm$2.57}} &47.30{\tiny$\pm$3.84} &39.25{\tiny$\pm$3.52} &30.35{\tiny$\pm$4.71} \\
        OpVuC
        &61.00{\tiny$\pm$1.39} &60.92{\tiny$\pm$1.54} &49.36{\tiny$\pm$1.14} &39.92{\tiny$\pm$2.66} &16.37{\tiny$\pm$3.74} &12.78{\tiny$\pm$2.90} &35.52{\tiny$\pm$1.82} &27.50{\tiny$\pm$2.59} &15.57{\tiny$\pm$2.14} &37.94{\tiny$\pm$0.79} &30.73{\tiny$\pm$1.58} &18.92{\tiny$\pm$1.94} \\
        \midrule
        PVC 
        &31.20{\tiny$\pm$4.19} &25.94{\tiny$\pm$3.99} &13.15{\tiny$\pm$2.69} &57.79{\tiny$\pm$2.38} &44.73{\tiny$\pm$2.92} &30.69{\tiny$\pm$4.82} &37.95{\tiny$\pm$0.85} &35.76{\tiny$\pm$0.19} &18.94{\tiny$\pm$0.32} &32.36{\tiny$\pm$0.48} &29.07{\tiny$\pm$0.18} &14.62{\tiny$\pm$0.11} \\
        MvCLN 
        &- &- &- &- &- &- &44.93{\tiny$\pm$0.54} &31.98{\tiny$\pm$0.70} &21.18{\tiny$\pm$0.65} &33.31{\tiny$\pm$0.45} &22.43{\tiny$\pm$1.11} &13.50{\tiny$\pm$0.65} \\
        SURE 
        &- &- &- &- &- &- &29.10{\tiny$\pm$0.18} &20.50{\tiny$\pm$0.52} &10.83{\tiny$\pm$0.39} &32.62{\tiny$\pm$0.20} &24.61{\tiny$\pm$0.75} &13.02{\tiny$\pm$0.28} \\
        CGDA 
        &51.80{\tiny$\pm$0.10} &49.54{\tiny$\pm$0.64} &27.93{\tiny$\pm$1.32} &50.23{\tiny$\pm$3.30} &33.29{\tiny$\pm$1.14} &14.56{\tiny$\pm$2.93} 
        &\multicolumn{3}{c|}{OOM} &\multicolumn{3}{c}{OOM} \\
        VITAL 
        &62.52{\tiny$\pm$3.92} &52.97{\tiny$\pm$2.62} &42.73{\tiny$\pm$3.38} &52.21{\tiny$\pm$4.48} &30.15{\tiny$\pm$3.64} &22.48{\tiny$\pm$5.07} &43.26{\tiny$\pm$3.69} &34.80{\tiny$\pm$3.23} &24.67{\tiny$\pm$3.24} &49.35{\tiny$\pm$3.20} &39.46{\tiny$\pm$2.63} &29.61{\tiny$\pm$3.32} \\
        CGCN 
        &52.38{\tiny$\pm$0.45} &42.07{\tiny$\pm$2.64} &30.42{\tiny$\pm$2.30} &60.27{\tiny$\pm$0.19} &29.39{\tiny$\pm$0.74} &29.47{\tiny$\pm$0.40} &48.93{\tiny$\pm$2.38} &37.73{\tiny$\pm$1.96} &26.24{\tiny$\pm$2.89} &47.12{\tiny$\pm$1.86} &36.22{\tiny$\pm$3.52} &22.11{\tiny$\pm$3.37} \\
        TCLPVC
        &66.68{\tiny$\pm$5.87} &60.16{\tiny$\pm$3.70} &47.59{\tiny$\pm$5.42} &59.98{\tiny$\pm$3.43} &42.98{\tiny$\pm$4.65} &36.77{\tiny$\pm$3.78} &41.72{\tiny$\pm$2.17} &36.88{\tiny$\pm$3.24} &26.21{\tiny$\pm$2.85} &42.80{\tiny$\pm$1.69} &38.94{\tiny$\pm$3.53} &25.74{\tiny$\pm$2.18} \\
        \rowcolor{green!10} ~ SMART ~ 
        &\textbf{68.03{\tiny$\pm$3.90}} &\textbf{62.31{\tiny$\pm$5.06}}  &51.55{\tiny$\pm$6.45}  &\textbf{69.37{\tiny$\pm$3.58}}  &\textbf{50.22{\tiny$\pm$3.20}}  &\textbf{33.60{\tiny$\pm$3.53}}  &\textbf{49.51{\tiny$\pm$1.74}}  &38.23{\tiny$\pm$2.90}  &28.24{\tiny$\pm$1.43}  &\textbf{54.70{\tiny$\pm$2.87}}  &\textbf{40.71{\tiny$\pm$1.91}}  &\textbf{31.97{\tiny$\pm$2.19}} \\
	\bottomrule[1.25pt]
    \end{tabular}
    }
    \label{tab:extreme_alignment}
\end{table*}
\subsubsection{\textbf{How Robust is SMART under Different Alignment Scenarios}}
By capturing the category semantics of both aligned and unaligned instance pairs, we enhance the robustness of our model, as illustrated in Fig. \ref{fig:Align_diffalign}. SMART exhibits exceptional robustness to varying alignment rates, outperforming both PVC and MVC methods across all metrics and datasets. Its minimal performance degradation, particularly on HandWritten, and low variability (tight standard deviations) demonstrate its reliability in challenging partial alignment scenarios. By effectively aligning view distributions and learning feature consistency from all data, our method proves to be a highly robust and adaptable solution for multi-view clustering, even under extreme misalignment conditions.

\begin{figure}[!t]
    \centering{
        \includegraphics[width=0.48\textwidth]{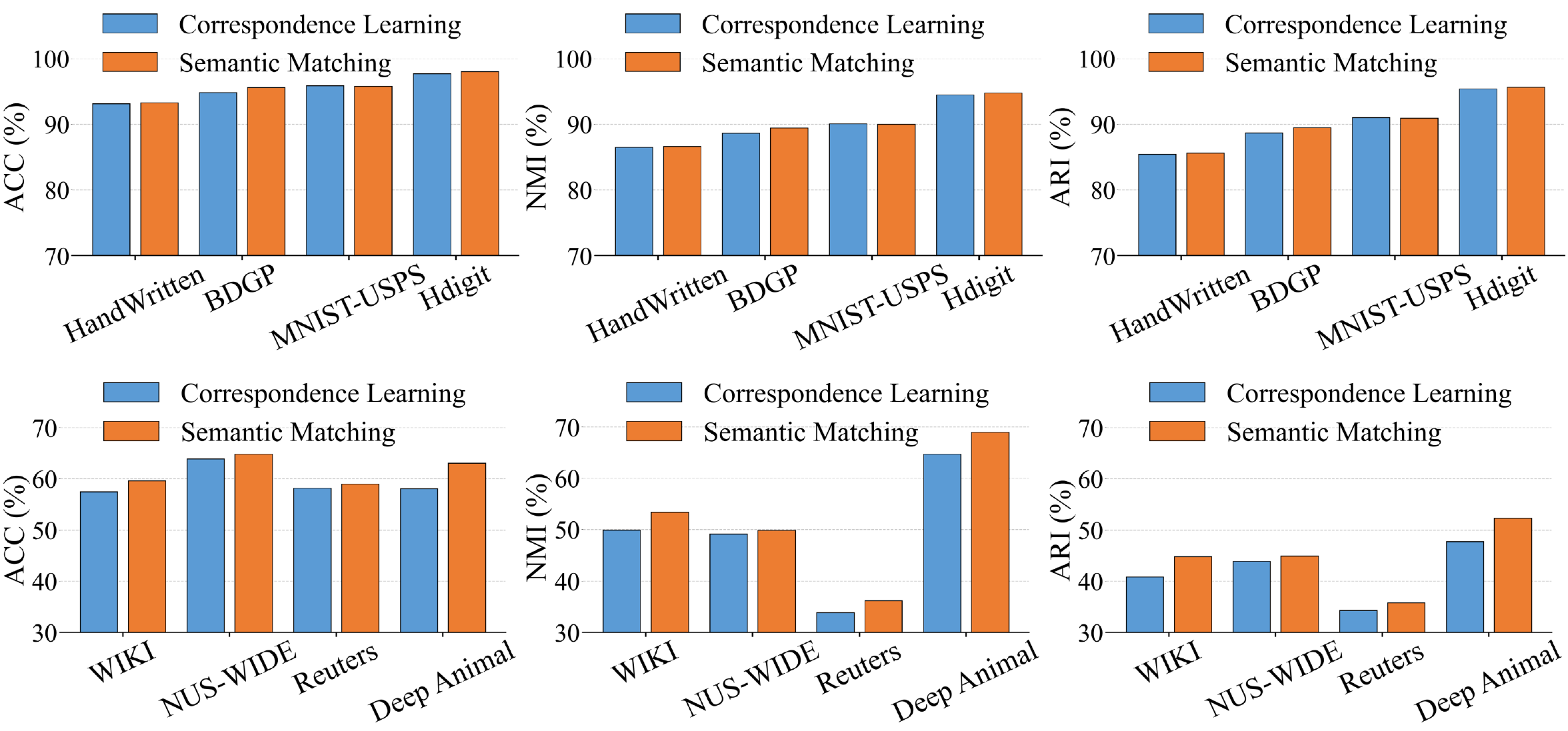}
        \caption{Performance comparison of the semantic matching against correspondence learning on eight datasets.}
    \label{fig:fusion}
    }
\end{figure}
\begin{figure*}[t]
    \centering
    \includegraphics[width=1.0\linewidth]{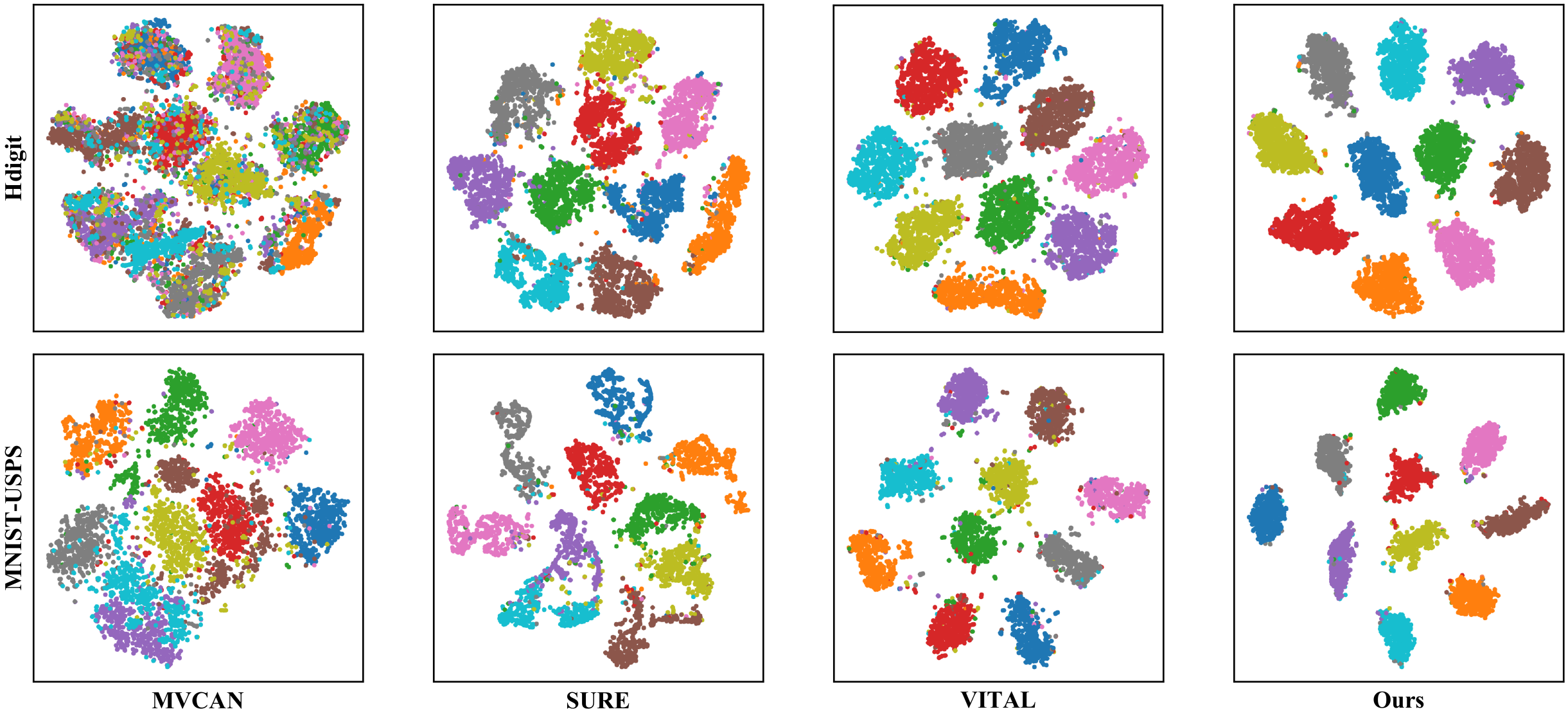}
    \caption{The t-SNE visualization on the Hdigit and MNIST-USPS datasets in partially view-aligned (50\%) scenario.}
    \label{fig:tsne_0.5}
\end{figure*}
\subsubsection{\textbf{Performance Comparison of the Semantic Matching against Correspondence Learning}}
To evaluate the efficacy of our proposed semantic matching feature fusion technique, we compare it against a correspondence-learning-based variant of SMART across eight datasets, including four single-modal multi-view datasets (HandWritten, MNIST-USPS, Reuters, and Hdigit) and four multi-modal multi-view datasets (BDGP, WIKI, NUS-WIDE, and Deep Animal). Following existing PVC methods \cite{2021MvCLN-yang, 2022SURE-yang}, we treat the samples with the smallest Euclidean distance as its correspondence in other views for realignment. As illustrated in Fig. \ref{fig:fusion}, the results clearly indicate that the clustering performance of semantic matching and correspondence learning is comparable on single-modal multi-view datasets, with no significant differences observed. In contrast, on multi-modal multi-view datasets such as BDGP, WIKI and Deep Animal, semantic matching demonstrates a clear advantage. This disparity arises from the fact that correspondence learning typically relies on distance metrics—such as the Euclidean distance—to perform instance- or category-level realignment across views. While this approach may work well when the views exhibit minimal feature heterogeneity, as in single-modal datasets like HandWritten and MNIST-USPS, they become less reliable in multi-modal scenarios due to significant feature distribution differences, leading to inaccurate realignment. In contrast, our approach mitigates cross-view feature heterogeneity via view distribution alignment and enhances representation learning through semantic matching contrastive learning. By extracting more generalized semantic consistency from reliable semantic neighbors, our method avoids the adverse effects of erroneous realignment, thereby improving the quality of learned embeddings. As a result, our semantic matching feature fusion technique demonstrates stronger and more robust representation capabilities across both single- and multi-modal multi-view datasets, leading to superior cluster assignments.

\subsubsection{\textbf{Performance under Extremely Limited Aligned Data}}
To further validate robustness, we conduct experiments under extremely low alignment rate (1\%), which is empirically equivalent to a fully unaligned condition, and compare SMART with both state-of-the-art PVC methods and fully unaligned MVC methods (i.e., MVC-UM \cite{2021MVC-UM-yu} and OpVuC \cite{2024OpVuC-dong}). The evaluation results in Table \ref{tab:extreme_alignment} clearly show that:
\begin{itemize}
    \item MVC-UM and OpVuC perform comparably to the best PVC methods on single-modal datasets (HandWritten, MNIST-USPS, and Hdigit), but exhibit a significant performance gap on the multi-modal dataset BDGP. This indicates that PVC methods still maintain advantages on datasets with strong view heterogeneity, as they can leverage limited cross-view correlations to enhance representation learning.
    \item All PVC methods experience a significant performance degradation compared to the 50\%/100\% alignment settings. Some methods, e.g., SURE and MvCLN, fail to produce valid clustering results due to the scarcity of aligned samples. This decline mainly stems from their inability to capture comprehensive category-level semantics when only a few aligned pairs are available, thereby impairing their discriminative capability.
    \item In contrast, SMART consistently achieves the best or comparable performance across all datasets under such condition. This advantage arises from SMART’s semantic matching contrastive learning, which effectively exploits unaligned data and alleviates the adverse effects of minimal alignment, enabling the model to extract more representative and semantically rich features.
\end{itemize}
\begin{figure*}[t]
    \centering
    \includegraphics[width=1.0\linewidth]{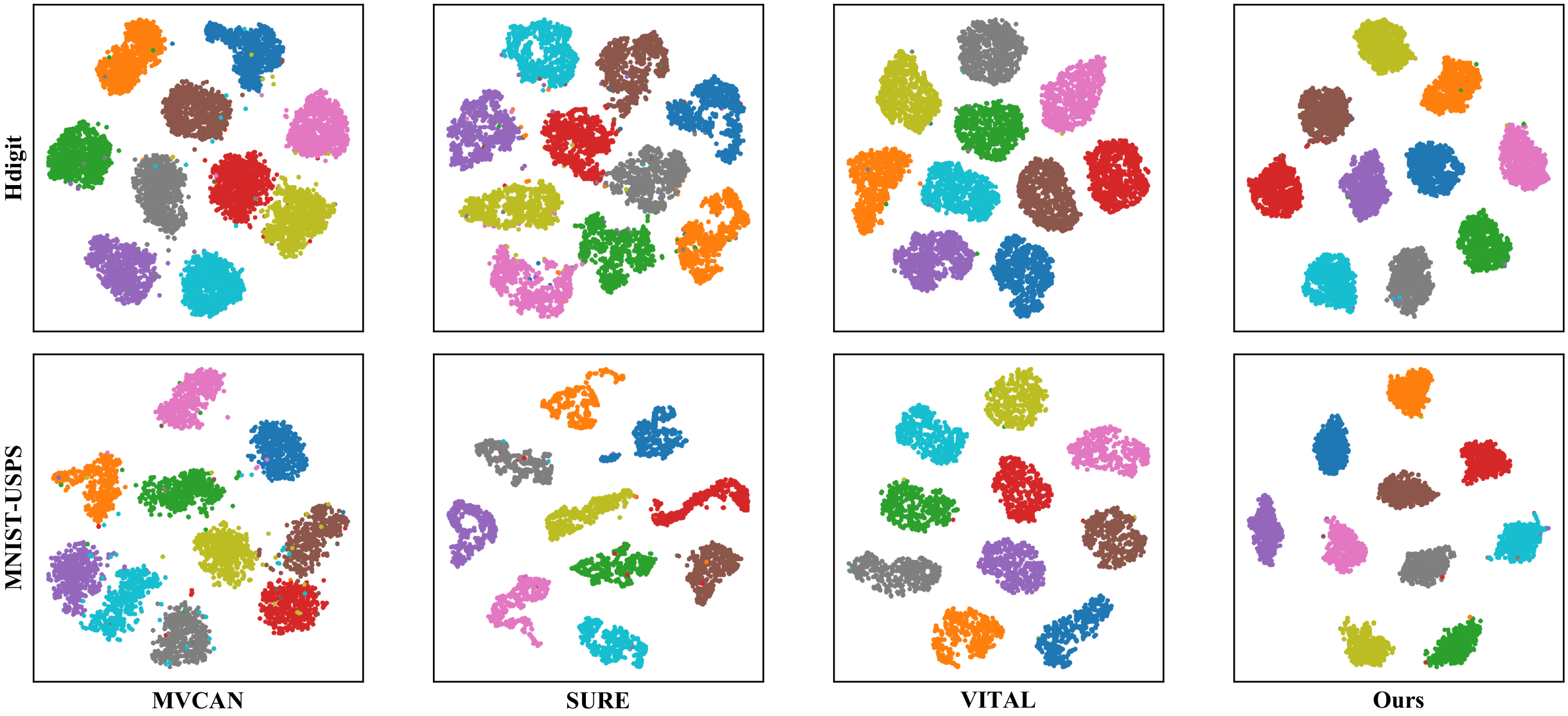}
    \caption{The t-SNE visualization on the Hdigit and MNIST-USPS datasets in fully view-aligned (100\%) scenario.}
    \label{fig:tsne_1.0}
\end{figure*}
\subsection{t-SNE Visualization}
\label{sub:t-SNE Visualization}
Fig. \ref{fig:tsne_0.5} and Fig. \ref{fig:tsne_1.0} illustrate the t-SNE \cite{2008tSNE-van} visualization of the fused representation $\textbf{H}$ on the Hdigit and MNIST-USPS datasets under the partially and fully view-aligned settings, respectively. Obviously, the PVC-specific methods outperform the latest generic MVC method MVCAN under $50\%$ alignment scenario. Among which, SURE and VITAL demonstrate notable improvements in cluster compactness and inter-class separation. While in $100\%$ alignment setting, the three baseline methods exhibit comparable cluster distributions. Nonetheless, our SMART achieves the most distinct clustering structure, characterized by highly compact intra-cluster distributions, well-separated inter-cluster boundaries, and fewer outliers. This advantage primarily stems from our view distribution alignment and semantic matching contrastive learning mechanisms, which enable the model to effectively capture more generalizable intra-cluster semantic consistency, and more distinguishable inter-cluster features. This underscores its superior ability to learn discriminative and robust representations from both partially and fully aligned data.

\begin{figure}[t]
    \centering{
        \includegraphics[width=0.48\textwidth]{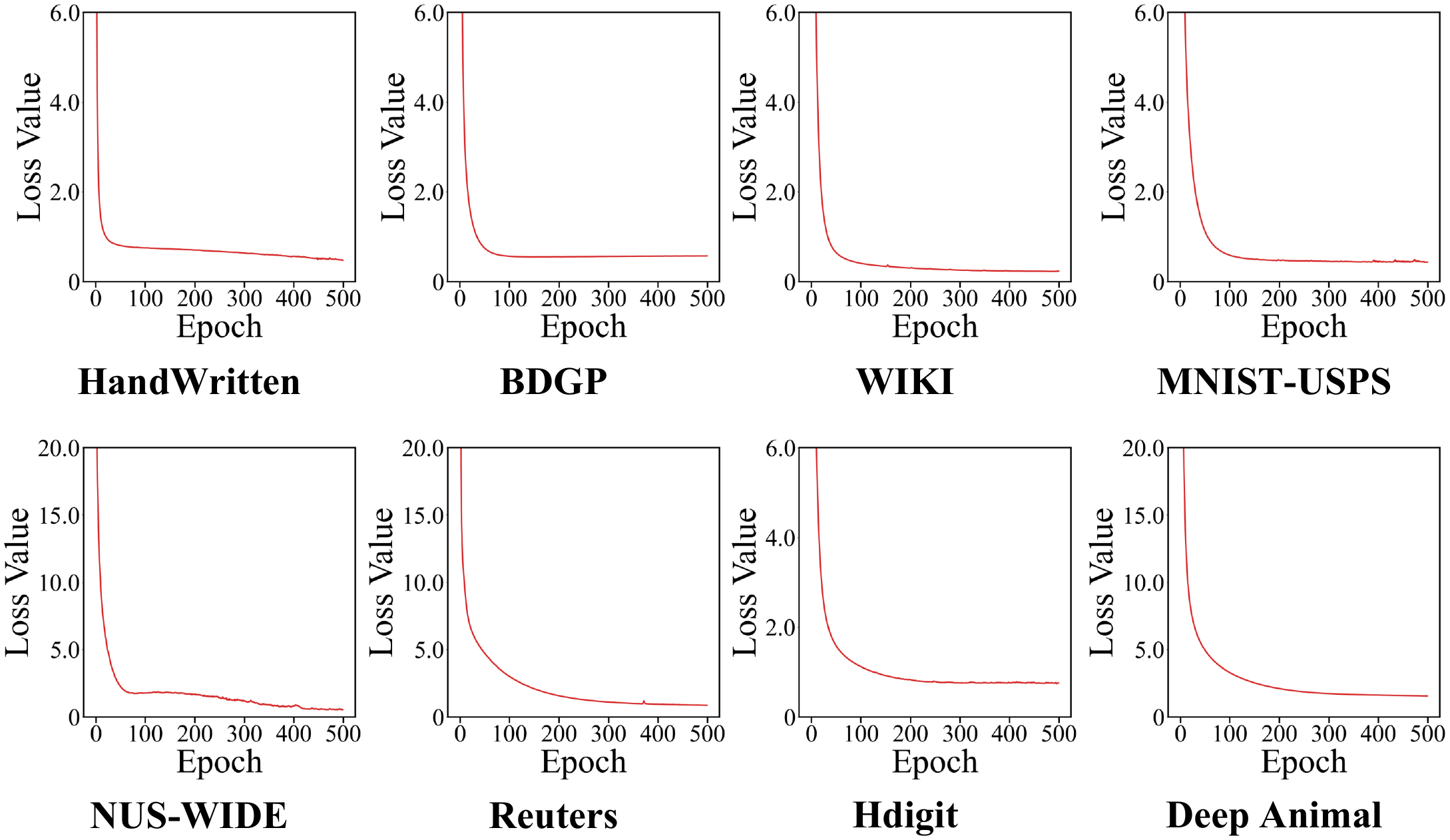}
        \caption{Loss curves on eight datasets.}
    \label{fig:loss_curves}
    }
\end{figure}
\subsection{Convergence Analysis}
To verify the convergence of the proposed optimization algorithm, we conducted convergence analysis on eight benchmark datasets: HandWritten, BDGP, WIKI, MNIST-USPS, NUS-WIDE, Reuters, Hdigit, and Deep Animal, as illustrated in Fig. \ref{fig:loss_curves}. The experimental results demonstrate that the algorithm exhibits strict convergence properties across all datasets: the loss function values monotonically decrease with increasing iterations and eventually stabilize. Specifically, the loss curves show no oscillations or divergence and typically reach the convergence threshold within a finite number of iterations, e.g. 200. This convergence behavior validates the theoretical soundness of the algorithm and confirms its efficient convergence performance.
\begin{figure}[!t]
    \centering{
        \includegraphics[width=0.48\textwidth]{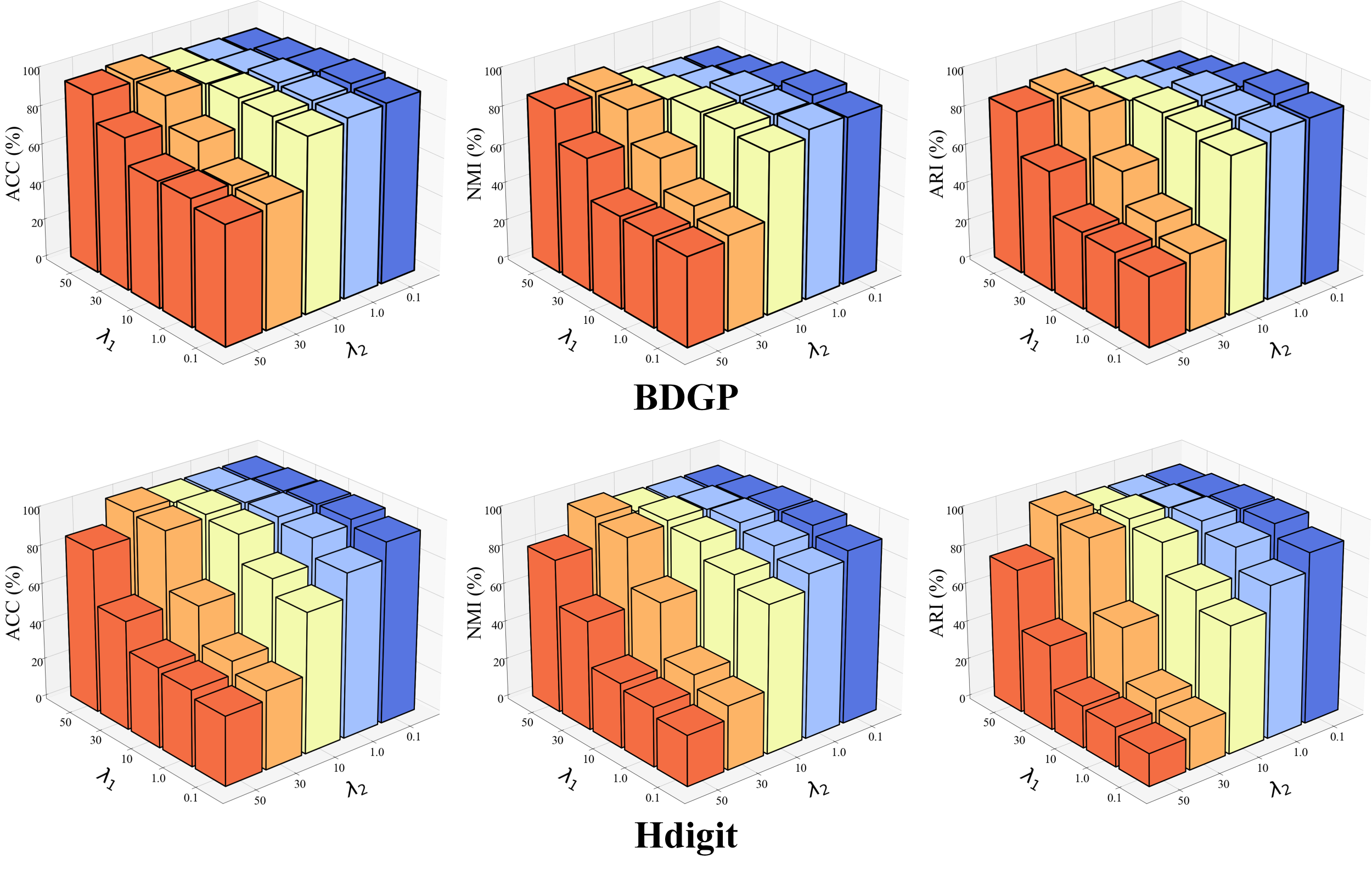}
        \caption{Joint sensitivity analysis of hyper-parameters $\lambda_1$ and $\lambda_2$ on the BDGP and Hdigit datasets.}
    \label{fig:lambdas}
    }
\end{figure}
\subsection{Hyper-Parameter Analysis}
\label{sec:parameter}
\subsubsection{\textbf{Trade-Off Factors $\lambda_1$ and $\lambda_2$}}
We analyze the sensitivity of the hyper-parameters $\lambda_1$ and $\lambda_2$ in the total loss (Eq. (\ref{Eq:loss_total})). Experiments are conducted on the BDGP and Hdigit datasets by varying $\lambda_1$ and $\lambda_2$ in the range [0.1, 1.0, 10.0, 30.0, 50.0], as shown in Fig. \ref{fig:lambdas}, while keeping other settings fixed. 
From the results, it is evident that when $\lambda_1$ is significantly smaller than $\lambda_2$—e.g., $\lambda_1 =0.1$, $\lambda_2=50$ —the model’s performance drops markedly. This decline suggests that an overly weak view distribution alignment term fails to adequately constrain the latent space, undermining the quality of the semantic guidance graph. Conversely, optimal performance is achieved when $\lambda_1$ is equal to or moderately larger than $\lambda_2$ (e.g., by up to two intervals, such as $\lambda_1=10$, $\lambda_2=1.0$), with both parameters within the range [0.1, 30]. 

This behavior indicates that appropriately increasing the strength of view distribution alignment ($\lambda_1$) enhances the quality of the semantic guidance graph, which in turn guides contrastive learning to produce more discriminative and higher-quality representations. The wide range of effective ($\lambda_1$, $\lambda_2$) combinations underscores the importance of their balance for clustering performance. Moreover, the consistent performance across this broad interval highlights the robustness of SMART to these hyper-parameters.

\subsubsection{\textbf{Latent Feature Dimension $d$}}
\begin{figure}[!t]
    \centering{
        \includegraphics[width=0.48\textwidth]{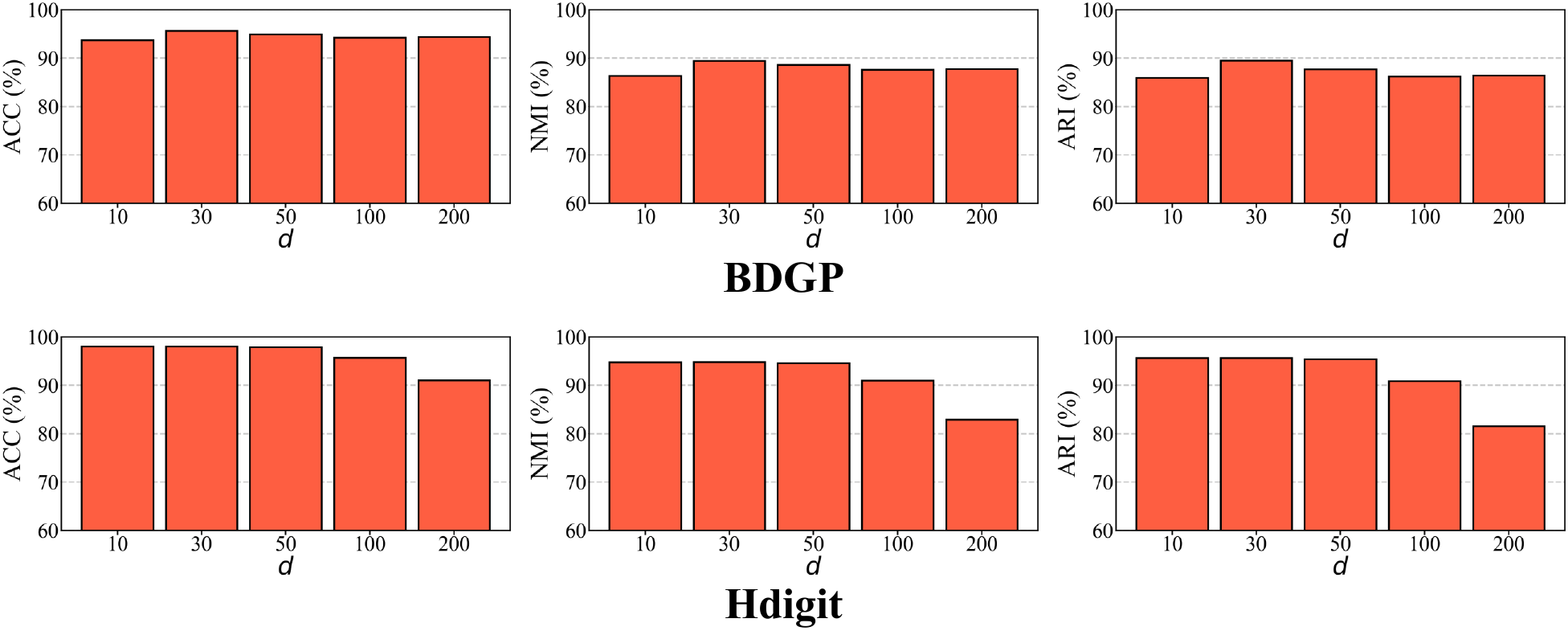}
        \caption{Sensitivity analysis of the latent dimension $d$ on the BDGP and Hdigit datasets.}
    \label{fig:dimension}
    }
\end{figure}

We further investigate the impact of the embedding dimension $d$ on clustering performance. We vary $d$ within the range $\{10, 30, 50, 100, 200 \}$ and present the clustering results in Fig. \ref{fig:dimension}. From the figure, we observe the following trends: On the BDGP dataset, the clustering performance first improves as $d$ increases from 10 to 30, followed by a slight decline as $d$ further increases up to 200. The optimal performance in terms of ACC, NMI, and ARI is achieved at $d=30$. On the Hdigit dataset, the clustering metrics remain at a high level when $d \in [10, 50]$, but all three evaluation metrics drop significantly when $d \ge 100$. These observations suggest that overly small dimensions (e.g., $d = 10$) may lead to insufficient representational capacity, degrading the quality of learned features. Conversely, an excessively large dimension (e.g., $d=200$) introduces noisy or redundant features, which can harm clustering performance. Therefore, selecting an appropriate embedding dimension is crucial for achieving high-quality cluster assignments.

\subsubsection{\textbf{Batch Size $B$}}
\begin{figure}[!t]
    \centering{
        \includegraphics[width=0.48\textwidth]{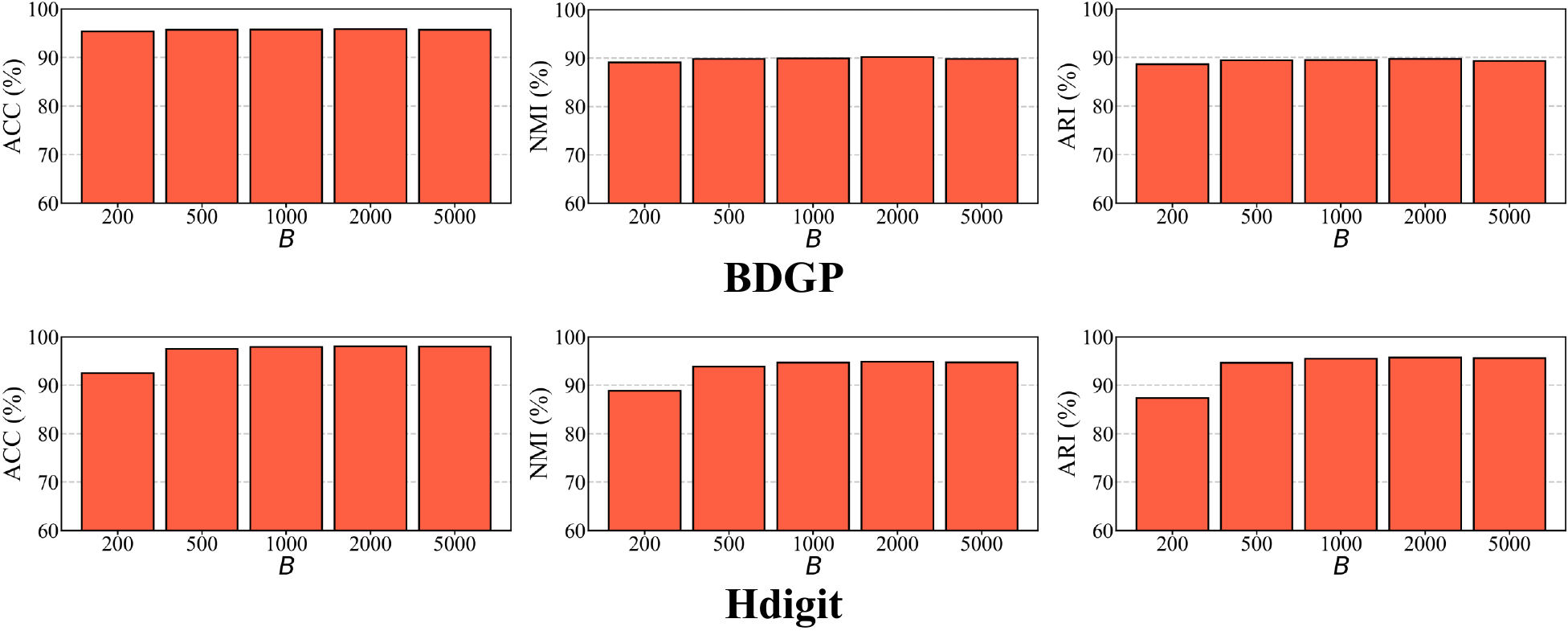}
        \caption{Sensitivity analysis of the batch size $B$ on the BDGP and Hdigit datasets.}
    \label{fig:batch_size}
    }
\end{figure}

During model training, we adopt a mini-batch optimization strategy with batch size $B$. To investigate its impact on model performance, we conduct a sensitivity analysis by varying $B$ across the range $\{200,500,1000,2000,5000\}$ on two representative benchmark datasets, BDGP and Hdigit. The results are illustrated in Fig. \ref{fig:batch_size}. On the BDGP dataset, it is evident that all three clustering metrics remain consistently stable across the entire range of batch sizes. For the Hdigit dataset, the model achieves optimal performance over a broad range of $[500,5000]$, with only a degradation is observed at the smallest batch size of 200. These findings demonstrate that our method exhibits strong robustness to the choice of batch size over a wide range, e.g., $[500,5000]$.

\section{Conclusion}
In this work, we propose a semantic matching contrastive learning method for PVC to address the limitations of existing approaches. Unlike existing PVC methods that primarily focus on learning view correspondences for unaligned data, our method performs view distribution alignment by aligning cross-view covariance matrices and constructs a reliable semantic graph to guide semantic matching contrastive learning, fully discovering the consistency and complementary semantic information from both aligned and unaligned data. Through the above optimization, this approach smoothly performs semantic matching to obtain meaningful and robust representations, which is free of cumbersome correspondence learning. This offers a novel solution for partially view-aligned clustering tasks, significantly enhancing the ability to capture reliable and comprehensive semantic relationships across views.

\section*{Acknowledgment}

This work was supported in part by Guangdong Basic and Applied Basic Research Foundation (Grant No. 2025A1515011692, 2023A1515030154), in part by Scientific Research Innovation Capability Support Project for Young Faculty (Grant No. ZYGXQNJSKYCXNLZCXM-H8), in part by National Natural Science Foundation of China (Project No.62106136, No. 62072189), in part by TCL Science
and Technology Innovation Fund (Grant No. 20231752), in
part by the Research Grants Council of the Hong Kong Special Administration Region (Grant No. CityU 11206622).

\bibliographystyle{IEEEtran}
\bibliography{reference}

\end{document}